\documentclass[final,5p,times,twocolumn]{elsarticle}

\usepackage{amssymb}

\usepackage{microtype}
\usepackage{graphicx}
\usepackage{float}
\usepackage{subcaption}
\usepackage{booktabs} 

\usepackage{amsmath} 
\usepackage{amssymb} 
\usepackage{hyperref}
\usepackage[linesnumbered,ruled,vlined]{algorithm2e}
\usepackage{amsmath}
\usepackage{amssymb}
\usepackage{algpseudocode}
\usepackage{xcolor}
\usepackage{hyperref}
\usepackage{multirow}
\usepackage{xcolor}
\definecolor{deepgreen}{rgb}{0,0.6,0}

\journal{Nuclear Physics B}

\begin{document}

\begin{frontmatter}



\title{MMSR: Symbolic Regression is a Multi-Modal Information Fusion Task}


\author[1,2, fn1]{Yanjie Li}
\ead{liyanjie@semi.ac.cn}
\author[1,2, fn1]{Jingyi Liu}
\ead{liujingyi@semi.ac.cn}


\author[1,2]{Min Wu \textsuperscript{*}}
\ead{wumin@semi.ac.cn}

\author[1,2]{Lina Yu \textsuperscript{*}}
\ead{yulina@semi.ac.cn}

\author[1,2,4]{Weijun Li \textsuperscript{*}}
\ead{wjli@semi.ac.cn}

\author[1,2]{Xin Ning}
\ead{ningxin@semi.ac.cn}

\author[1,4]{Wenqiang Li}
\ead{liwenqiang@semi.ac.cn}
\author[1]{Meilan Hao}
\ead{mlhao@semi.ac.cn}

\author[1,2]{Yusong Deng}
\ead{ysdeng@semi.ac.cn}

\author[1,2]{Shu Wei}
\ead{weisu@semi.ac.cn}

\cortext[cor1]{Corresponding Authors}
\fntext[fn1]{The first two authors contributed equally to this work.}

\affiliation[1]{organization={AnnLab, Institute of Semiconductor, Chinese Academy of Sciences},
            city={Beijing},
            postcode={100083}, 
            country={China}}
\affiliation[2]{organization={Center of Materials Science and Optoelectronics Engineering $\&$ School of Integrated Circuits, University of Chinese Academy of Sciences},
            city={Beijing},
            postcode={101408}, 
            country={China}}

\affiliation[4]{organization={School of Electronic, Electrical and Communication Engineering,
University of Chinese Academy of Science},
            city={Beijing},
            postcode={101408}, 
            country={China}}


\begin{abstract}
Mathematical formulas are the crystallization of human wisdom in exploring the laws of nature for thousands of years. Describing the complex laws of nature with a concise mathematical formula is a constant pursuit of scientists and a great challenge for artificial intelligence. This field is called symbolic regression  (SR). Symbolic regression was originally formulated as a combinatorial optimization problem, and Genetic Programming (GP) and Reinforcement Learning algorithms were used to solve it. However, GP is sensitive to hyperparameters, and these two types of algorithms are inefficient. To solve this problem, researchers treat the mapping from data to expressions as a translation problem. And the corresponding large-scale pre-trained model is introduced. However, the data and expression skeletons do not have very clear word correspondences as the two languages do. Instead, they are more like two modalities (e.g., image and text). Therefore, in this paper, we proposed MMSR. The SR problem is solved as a pure multi-modal problem, and contrastive learning is also introduced in the training process for modal alignment to facilitate later modal feature fusion. It is worth noting that to better promote the modal feature fusion, we adopt the strategy of training contrastive learning loss and other losses at the same time, which only needs one-step training, instead of training contrastive learning loss first and then training other losses. Because our experiments prove training together can make the feature extraction module and feature fusion module wearing-in better. Experimental results show that compared with multiple large-scale pre-training baselines, MMSR achieves the most advanced results on multiple mainstream datasets including SRBench. Our code is open source at \href{https://github.com/1716757342/MMSR}{https://github.com/1716757342/MMSR}.
\end{abstract}

\begin{graphicalabstract}
\end{graphicalabstract}

\begin{highlights}
     \item Propose a new multi-modal-based approach for symbolic regression.
    \item Achieve an effective modality alignment mechanism utilizing contrastive learning.
    \item Design a novel joint training strategy based on multi-loss fuction.
    \item Our code is open source at \href{https://github.com/1716757342/MMSR}{https://github.com/1716757342/MMSR}.

\end{highlights}


\end{frontmatter}


\section{Introduction}
Symbolic Regression (SR) aims to identify the explicit mathematical expression underlying the observed data. Formally, for a given dataset $D=\{X,y\}$, SR seeks a function $f$ that satisfies $y=f(X)$, where $X\in \mathbb{R}^{n\times d}$, $y\in \mathbb{R}^n$, $d,n$ are the dimension of variable and number of data points, respectively. $f$ is composed of several basic primitive operators such as $+,-,\times,\div$, etc. SR has been applied in many fields, for example, material sciences\cite{wang_wagner_rondinelli_2019}, physical law exploration\cite{doi:10.1126/sciadv.aay2631}\cite{NEURIPS2020_33a854e2}\cite{doi:10.1126/science.1165893}, etc.

In symbolic regression, expressions are usually represented as binary trees. Genetic Programming Symbolic Regression (GPSR)\cite{gpsr1}\cite{gpsr2}\cite{gpsr3}\cite{gpsr4}\cite{gpsr5}\cite{gpsr6} has dominated in the past. GPSR uses the pre-order traversal of expression and acts as genes, and genetic operations such as mutation and crossover are used to produce descendants. Constants are randomly generated or optimized with the Particle Swarm Optimization (PSO)\cite{jain2022overview}. Although they have shown distinguished performances in several public benchmarks, they are sensitive to hyperparameters \cite{DBLP:journals/corr/abs-1912-04871} and the complexity of the learned expression tends to be high with the epoch increased.

With the huge success gained by deep learning in multiple domains \cite{radford2018improving}, some works apply the deep learning technique in SR. Equation learner (EQL) \cite{9180100} first substitutes the activation function of a fully connected feedforward neural network and refines the expression with Lasso. However, EQL is hard to tackle the division operation because it would generate exaggerated gradients in backpropagation. Therefore, Deep Symbolic Regression (DSR) \cite{DBLP:journals/corr/abs-1912-04871} utilizes an RNN to emit the skeleton of expressions, constants are represented with a constant placeholder and optimized later by the numerical optimization algorithm. The risk-seeking policy gradients are used to update the RNN. To step further, Neural-Guided Genetic Programming (DSO, NGGP) \cite{DBLP:journals/corr/abs-2111-00053} incorporates genetic programming into the search, RNN provides a better population for genetic programming algorithms. By doing so, DSO achieves remarkable results in SRBench \cite{la2021contemporary}, Since the efficiency of reinforcement learning is lower than that of numerical optimization, MetaSymNet\cite{li2023metasymnet} skillfully uses numerical optimization problems to solve the combinatorial optimization problem of symbolic regression.
\begin{figure*}[ht]
\begin{center}
\centerline{\includegraphics[width=0.86\linewidth]{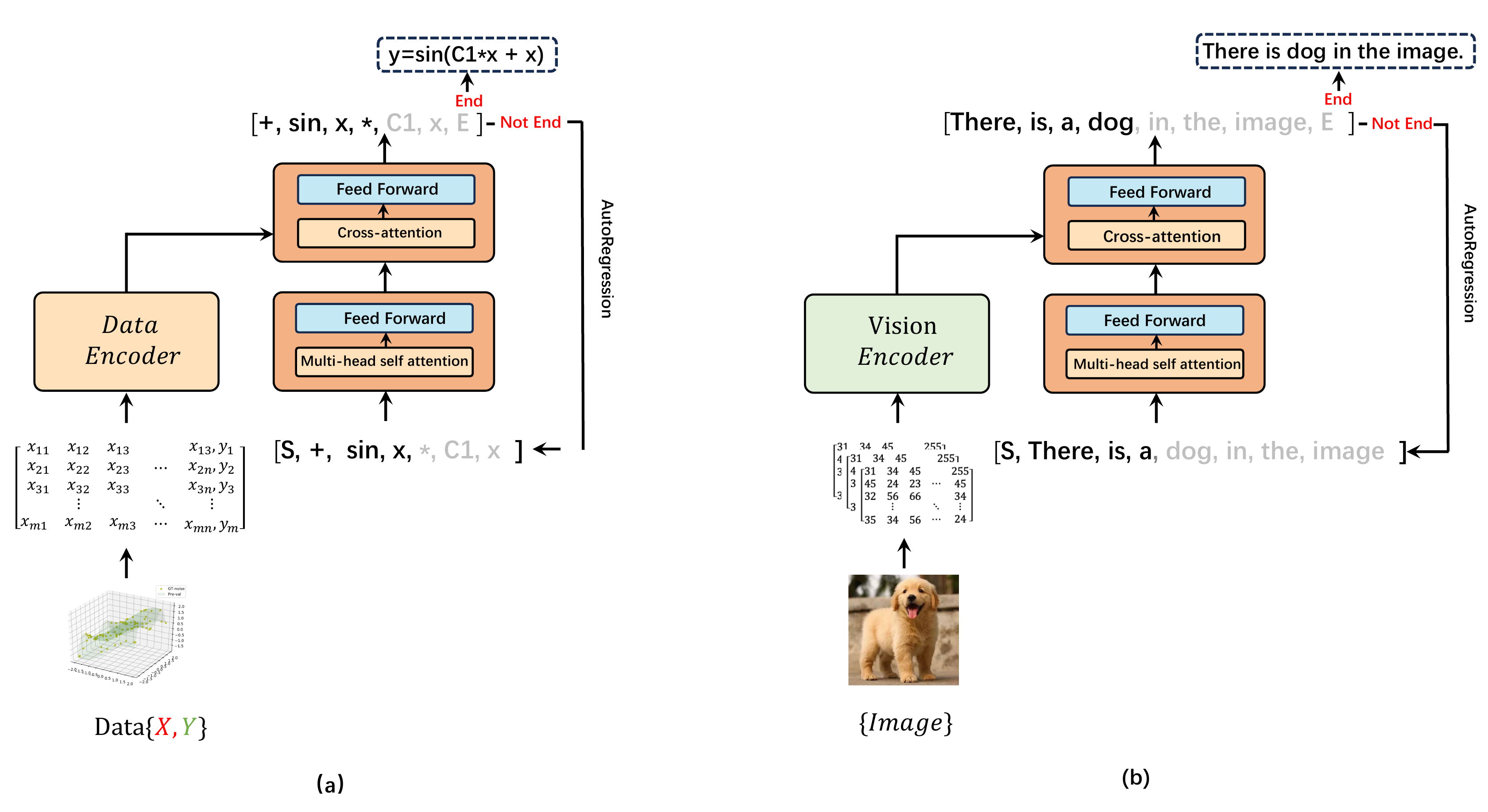}}
\caption{Multi-model-based Symbolic Regression task versus multi-model-based Image Caption task. (a) Multi-model-based Symbolic regression task. In symbolic Regression task, an expression is expected to get from a set of input data [X,Y]. It can be considered as a multi-modal task, where the expression and the data are two different modalities. (b) Multi-model-based Image Caption task. The text modality is output to caption an image. }
\label{fig:VQA}
\end{center}
\end{figure*}

The above-stated methods belong to the end-to-end model which trains a new model for each problem. This type of training strategy makes their search time long. Plus, the model can't make full use of the search experiences from other searches. Therefore, the large-scale pre-trained model is proposed to improve the search speed. Normally, the large-scale pre-trained models \cite{DBLP:journals/corr/abs-2106-14131}\cite{DBLP:conf/icml/BiggioBNLP21}\cite{DBLP:journals/corr/abs-2205-15764} 
treat the symbolic regression problem as a language translation problem and train with enormous self-generated data which contains the pairwise of sampled data points and corresponding mathematical expressions skeletons. The neural network has an encoder-decoder structure, the encoder maps the input data points into data features and the decoder uses the feature to generate the pre-order traversal of expressions. The Beam Search \cite{kumar2013beam} technique is applied to produce multiple candidate skeletons. Then, Broyden-Fletcher-Goldfarb-Shanno (BFGS) \cite{liu1989limited} is used to optimize the constants of candidate skeletons. Finally, the best-performed expression is selected as the final result.

The large-scale pre-trained model gained fast search speed because once the model finished training, we only needed a single forward pass to get the expressions. However, they ignore the different data formats of the input and output of SR. That is, previous methods treat the SR as a machine translation task: translate the data points into corresponding expressions. However, it is clear that the input and output of SR are very different from the two languages in translation, and they do not have obvious word correspondence like languages. They are more like two different modalities (like image and text). Treating an SR question as a multi-modal task and cross-referenced this process with an image caption task is illustrated in Fig. \ref{fig:VQA}.
Therefore, it is perhaps more reasonable to think of SR as a multi-modal task.

To tackle the problems of two different modalities, we use a modal alignment module in our model to make the encoded data feature close to the skeleton feature. We name our model as MMSR. Specifically, we use the SetTransformer \cite{pmlr-v97-lee19d} as a data encoder for it presents permutation invariance. And, decodes the skeleton of expression with a causal masking transformer decoder. Unlike standard decoder transformers, MMSR omits cross-attention in the first half of the decoder layers to encode unimodal skeleton representations, and cascades the rest of the decoder layers, cross-attending to the data encoder for multi-modal data-skeleton representations. 
We use contrastive learning \cite{chuang2020debiased} to align the modal features, so that the matching data features and skeleton features are as close as possible in the feature space, and the non-matching features are as far away from each other as possible.
Additionally, we train the model with two extra losses. First, the cross-entropy loss is used for predicting the token of expression skeletons. Second, one of the limitations of pre-train models is that they focus on the supervised pretraining goals borrowed from symbol generation, i.e., they are trained solely with the token-level cross-entropy (CE) loss, which can result in equations that may exhibit high token-level similarities but are suboptimal with respect to equation-specific objectives such as fitting accuracy\cite{shojaee2024transformer}. To solve this problem, we directly predict the numerical value of the constants and use MSE to measure the loss between the predicted constants and the true constants in the skeletons. The main contributions are summarized as follows:
\begin{itemize}
    \item Propose a new multi-modal-based approach for symbolic regression. The idea of solving a symbolic regression problem via modal alignment and modal feature fusion significantly enhances the ability to accurately obtain concise expressions from data.
    \item Achieve an effective modality alignment mechanism utilizing contrastive learning. In this mechanism, deep-level feature representations are derived by contrasting positive and negative samples, and subsequently mapped them into a feature space that matches data and expressions. It promotes an enhanced understanding and differentiation of the differences within the data, thereby significantly improving the model's comprehension capability regarding data and expressions.
    \item Design a novel joint training strategy based on multi-loss fuction. This strategy can adaptively achieve the optimal balance of expression skeleton, modal features, and data fitting accuracy under the collaborative constraints of cross-entropy loss ($\mathcal{L_{CE}}$), mean squared error (MSE) loss ( $\mathcal{L_{MSE}}$ ), and contrastive loss ( $\mathcal{L_{CON}}$ ), without over-relying on a single loss function.

\end{itemize}

The rest of the paper is organized as follows: Section \ref{related_works} introduces the relevant research of our work, Section \ref{method} describes the model structure, Section \ref{experiments} lists the experiment settings and displays the experiment results, experiment analysis is also included. Finally, Section \ref{conclusion} gives a discussion and conclusion of our work.

\section{Related Works}
\label{related_works}
\subsection{Multimodal}
Recently, models such as Contrastive Language-Image Pre-Training (CLIP) \cite{CLIP} and ALIGN \cite{ALIGN} performed pre-train on noisy images and text pairs from the web using contrastive loss. Contrastive loss is one of the most effective feature learning methods\cite{he2020momentum}\cite{chen2020simple}\cite{li2020prototypical}\cite{li2020mopro}. They achieve remarkable performance on image-text retrieval tasks but lack the ability to model more complex interactions between image and text for other V+L tasks \cite{kim2021vilt}. (e.g. VQA\cite{antol2015vqa}).
Subsequent investigations \cite{wang2021simvlm} \cite{wang2022ofa} \cite{piergiovanni2022answer} have introduced encoder-decoder frameworks trained utilizing generative loss functions, which demonstrate robust performance across vision-language benchmarks. Concurrently, the visual encoders within these models maintain competitive accuracy in image classification tasks.
Studies \cite{singh2022flava} \cite{li2021align} \cite{li2022blip}\cite{chen2023pali}\cite{liu2024visual} have investigated the unification of image and text representations, which require multiple pretraining stages of unimodal and multi-modal modules to achieve high-performance levels. For example, ALBEF \cite{li2021align}  employs a dual-encoder architecture that integrates contrastive loss with Masked Language Modeling (MLM) to enhance learning efficiency. CoCa\cite{yu2205coca} focuses on training an image-text foundation model from scratch in a single pretraining stage to unify these approaches. This is simpler and more efficient to train. BEITv3\cite{beit3} regards images as a kind of language, and maps images into the language space through a mapping layer, and then sends them into the GPT large model together with the encoded text features. LLava\cite{llava} is an open-source multi-modal large model that uses images as a language. First, the big oracle model is frozen, and pre-trained, and the picture features are aligned with the text features to turn the picture into a 'language' that the large language model(LLM)\cite{llm1}\cite{llm2}\cite{llm3}\cite{llm4}\cite{llm5} can understand, and then concatenated with the text features and sent to the large language model. 
\subsection{Deep Symbolic Regression}
Before the explosion of deep symbolic regression methods, Genetic Programming Symbolic Regression (GPSR) \cite{10.1145/2576768.2598291}\cite{McConaghy2011}\cite{7473913}\cite{9234005} was the main strategy for solving symbolic regression tasks. GPSR encodes the pre-order traversal of mathematical expression as the gene type, by randomly initializing a population of individuals, GPSR generates the candidate set of expressions. Then, it uses a fitness function to evaluate the fitting results, and genetic operations such as mutation and crossover are applied to produce new generations. The algorithm iterates until reaches the expected fitness value or the pre-defined maximum iteration epoch. Although GPSR could perform well in public benchmarks, it is reported to be sensitive to hyperparameters \cite{DBLP:journals/corr/abs-1912-04871}. Furthermore, the GP-based method lacks an effective way to determine the constant value in the expression. RSRM\cite{xureinforcement} skillfully combines GP, Monte Carlo Tree Search (MCTS), and Double Q-learning. MSDB block is introduced to define new operators heuristically, which greatly improves the running efficiency of the algorithm.
With the huge success achieved by deep learning in various domains, EQL revised the activation functions of fully connected neural networks to primitives (sin, cos, etc.) and obtained the skeleton and constants through Lasso \cite{https://doi.org/10.1111/j.2517-6161.1996.tb02080.x} technique. However, the nontraditional activation functions cause troubles to the backpropagation process: the division operator would cause infinite gradients thus making the weights updating hard to continue. Moreover, the primitive sin, cos are periodical, and log, sqrt are not differentiable everywhere. DSR \cite{DBLP:journals/corr/abs-1912-04871} takes another learning strategy: it learns the skeleton first and optimizes the constants later. DSR uses an RNN to emit the probability distribution of the pre-order traversal, and autoregressively sample expressions according to the distribution. For gradient information lost in the sampling process, reinforcement learning is applied to update the RNN. Specifically, the risk-seeking policy gradients are used. Furthermore, to utilize the strong ability to generate new performed well expressions of GPSR, DSO \cite{DBLP:journals/corr/abs-2111-00053} incorporating the DSR and GPSR method to achieve state-of-the-art performance in SRBench. The RNN could provide a better population for GPSR to produce new descendants.

DSO is a type of learning-from-scratch method, it needs to train a new model for every problem. This training strategy comes with a huge search time cost. Thus, many learning-with-experience methods are emerged. They usually employ a data encoder to receive the data points as inputs and output the expression encoding as the label to train a large-scale model. When a new problem needs to be handled, the learning-with-experience method just implements a forward propagation to obtain the expressions thus being time-effective. SymbolicGPT \cite{DBLP:journals/corr/abs-2106-14131} treats the problem as a natural language problem, it uses symbolic-level encoding as the label of a set of data points. For example, the expression $2sin(x+y)$ is represented as text $['2','s','i','n','(','x','+','y',')']$. This type of encoding forces the model to learn the syntax and semantics of the primitives thus increasing the learning difficulty. To free the model from learning the syntax and semantics meaning of primitives, NeSymReS \cite{DBLP:conf/icml/BiggioBNLP21} utilizes the pre-order traversal of binary tree to encode the expression. The constants in the expression are replaced with a placeholder ``$c$'' and optimized later after the skeleton is sampled according to the probability distribution outputted by the network. The pre-trained large-scale models have fast inference speed to obtain the predicted expression when confronted with new SR problems.
The critical challenge of the SR task lies in two aspects: determining the skeleton and optimizing the constants. The above-stated large-scale pre-trained model could do well in predicting the skeleton, however, they normally use BFGS to optimize the constants which are sensitive to initial guesses. Therefore, to determine the constants accurately, based on the NeSymRe framework, Symformer \cite{DBLP:journals/corr/abs-2205-15764} and End2End \cite{kamienny2022end} design two types of constant encoding strategies to teach the pre-trained model the exact value of constants. They perform with outstanding fitting accuracy compared to NeSymReS.

Beyond the constant optimizing problem, Li \textit{et al} \cite{li2022transformer} uses the Intra-class contrastive loss to train a better data encoder to tackle the ill-posed problem in SR. UDSR \cite{landajuela2022unified}, SNR \cite{liu2023snr}, and DGSR\cite{holt2023deep} merged multiple learning strategies to give the pre-trained model a revised mechanism. Moreover, the Monte Carlo Tree Search \cite{browne2012survey} is incorporated into the pre-trained model to achieve better performances.e.g.\cite{li2024discovering}. However, all the works view the SR task as a machine translation task that considers the input and output of the pre-trained model as the same data type. They ignore the fact that the data points are set data and the expression label is text. Thus, our work designs new architecture to solve the modality problem. SNIP\cite{meidani2023snip} employs a two-stage strategy to train a symbolic regression model. SNIP pre-trains an encoder using contrastive learning and then trains a decoder. Since the two-stage training strategy is not conducive to feature fusion, SNIP also needs to introduce the LSO method to work well. TPSR\cite{shojaee2024transformer} combines the large-scale pre-trained model and MCTS and uses the pre-trained model as a policy network to guide MCTS to search. The search efficiency of MCTS is improved.

\section{Method}
\label{method}
\subsection{Expression Representation}
\begin{figure}[ht]
\begin{center}
\centerline{\includegraphics[width=0.8\linewidth]{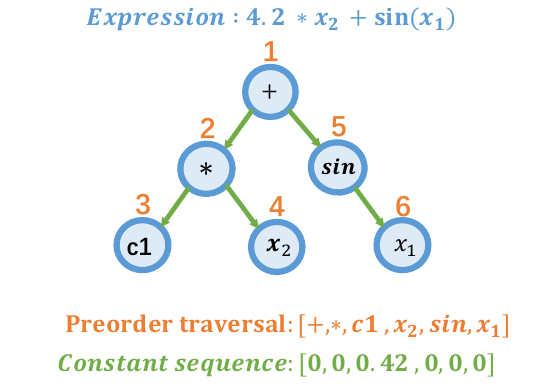}}
\caption{Expression binary tree. A binary tree representation of the expression $y = 4.2*x_2 + \sin(x_1)$ and a preorder traversal (skeleton) of its binary tree.}
\label{fig11}
\end{center}
\end{figure}

We represent the expression in the form of a binary tree. For example, for the expression $y = 4.2*x_1 + sin(x_2)$, the form of the expression binary tree is shown in Fig. \ref{fig11}, and we expand it in preorder traversal order to obtain a symbol sequence (expression skeleton). It can be thought of as a `sentence' where each symbol is considered a word. So we can cleverly use generative models to sequentially generate expression skeletons, just like large language models generate sentences.

\begin{figure*}[ht]
\begin{center}
\centerline{\includegraphics[width=0.98\linewidth]{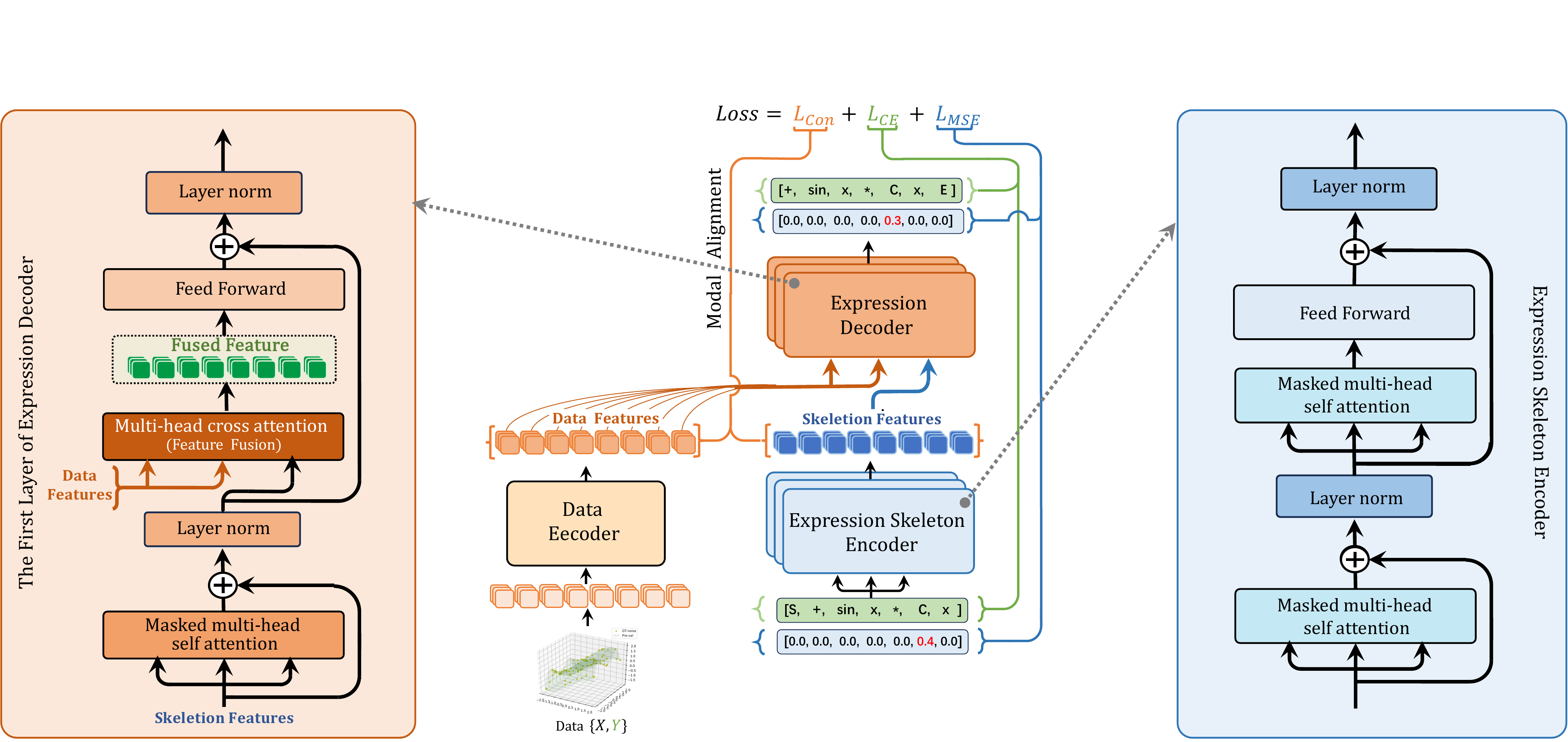}}
\caption{The framework of MMSR model. Among them, the fig. (b) shows the overall algorithm flowchart of MMSR. The fig. (a) shows the structure of the skeleton encoder of MMSR, and the fig. (c) shows the structure of the decoder, note that in the decoder, we use the cross-attention mechanism in the Multi-head cross-attention module for information fusion.}
\label{framework}
\end{center}
\vskip -0.2in
\end{figure*}

\subsection{Architecture of MMSR}
As depicted in Fig. \ref{framework}, and Algorithm \ref{alg1} the MMSR consists of two transformer-based encoders, each tailored for learning the symbolic or data representations of mathematical functions. Except that, a decoder receives the encoded numeric representations to decode the features to mathematical expressions. Additionally, an alignment module is designed to match the feature representations of numeric inputs and symbolic inputs. The contrastive learning \cite{chuang2020debiased} is used to make the distance between the data and expression skeleton features that are derived from the same expression closer. During training, MMSR receives synthetically created symbolic equations and their associated numeric data as inputs to the symbolic encoder and numeric encoder, respectively. In total, MMSR is trained on approximately 10 million synthetic paired (numeric, symbolic) examples.

\begin{algorithm}
\caption{MMSR}
\label{alg1}
\KwIn{data: $[X,y]$, Expression Skeleton: $S$, Constants: $C$, Epoch: $E$, Number of batch: $N$, $\lambda_{1}$, $\lambda_{2}$, $\lambda_{3}$.}
\Repeat{Arrival of termination condition}{
  Initialize $MMSR = M_{\theta}$.\\
  \For{$i=1$ \KwTo $E$}{
    \For{$j=1$ \KwTo $N$}{
      $[X_j, y_j]$, [$S_j$] = GetBatch($[X,y]$, [$S$])\\
      $F_{Data} = \text{SetTransformer}([X_j, y_j])$\\
      $F_{Skeleton} = \text{SkeletonEncoder}([S_j])$\\
      $S_{pred}, C_{pred} = \text{ExpressionDecoder}(F_{Data}, F_{Skeleton})$\\
      $\mathcal{L_{CON}} = \text{ContrastiveLoss}(F_{Data}, F_{Skeleton})$\\
      $\mathcal{L_{CE}} = \text{CrossEntropy}(S_i, S_{pred})$\\
      $\mathcal{L_{MSE}} = \text{MSE}(C_{i}, C_{pred})$\\
      $\mathcal{L} = \lambda_{1}\mathcal{L_{CE}} + \lambda_{2}\mathcal{L_{MSE}} + \lambda_{3}\mathcal{L_{CON}}$\\
    }
  Compute the gradient $\nabla_{\theta} \mathcal{L}$ and use it to update $\theta$.\\
  }
}
\end{algorithm}

\subsubsection{Data Encoder}
Considering the permutation invariance of the data feature: it should maintain the same feature with the order change of data input. We thus applied SetTransformer \cite{pmlr-v97-lee19d} to be our data encoder. The input to the encoder consists of the data points $\mathcal{D}=\{X,y\}\in\mathbb{R}^{n\times d}$, which are first passed through a trainable affine layer to project them into a latent space $h_n\in\mathbb{R}^{d_h}$. The resulting vectors are then passed through several induced set attention blocks \cite{pmlr-v97-lee19d}, which are several cross-attention layers. First, cross-attention uses a set of trainable vectors as the queries and the input features as keys and values. Its output is used as the keys and values for the second cross attention, and the original input vectors are used as the queries. After these cross-attention layers, we add a dropout layer \cite{2014Dropout}. In the end, we compute cross attention between a set of trainable vectors (queries) to fix the size of the output such that it does not depend on the number of input points.

\subsubsection{Expression Skeleton Encoder}
Symbolic expression by [+, -, $\times$, $\div$, sin, cos, exp, log \\ sqre, C-1, C-2,...,C4, C5, $x_1$, $x_2$,... ] and other basic operators, where [C-5,...,C5] represents a constant placeholder. Each expression can be represented by an expression binary tree, and we can get a symbol sequence by expanding the expression binary tree in preorder, We call this the expression skeleton. Furthermore, we treat each symbol as a $token$ and then embed it. 

In particular, to solve the problem that expressions may exhibit high token-level similarities but are suboptimal for equation-specific objectives such as fitting accuracy, which is caused by only using the cross-entropy loss between the generated sequence and the real sequence, we encode the expression symbol sequence and the constant sequence separately. For example, for the expression $0.021 * sin(x) + 12.2$, we would have the skeleton [+, $\times$, sin, x, C-1, C2]. And constant sequence $[0,0,0,0.21,0.122]$. It is worth noting that we replace constants with special symbols. The constants are encoded using a scientific-like notation where a constant C is represented as a tuple of the exponent $c_e$ and the mantissa $c_m$,  $C=c_m10^{c_e}$. Where $c_e$ belongs to integers between [-5,5] and $c_m$ belongs to [-1,-0.1] $\vee$ [0.1,1]. For example, for the constant 0.021 above, we use the placeholder C-1 in the skeleton, where $c_m$ is equal to 0.21 and $c_e$ is equal to -1, so $0.021=0.21*10^{-1}$.

When calculating the loss function during training, we will calculate the cross-entropy loss for the predicted and true symbol sequences, and the numerical MSE loss for the predicted and true constant sequences. Note that when calculating the loss, we pad both the skeleton of the expression and the corresponding constants sequence to the maximum allowed length N, where the letter P is used for symbols and 0.0 is used for constants.
\begin{figure*}[htp]
\centering
\hspace{0.1cm}
\includegraphics[width=180mm]{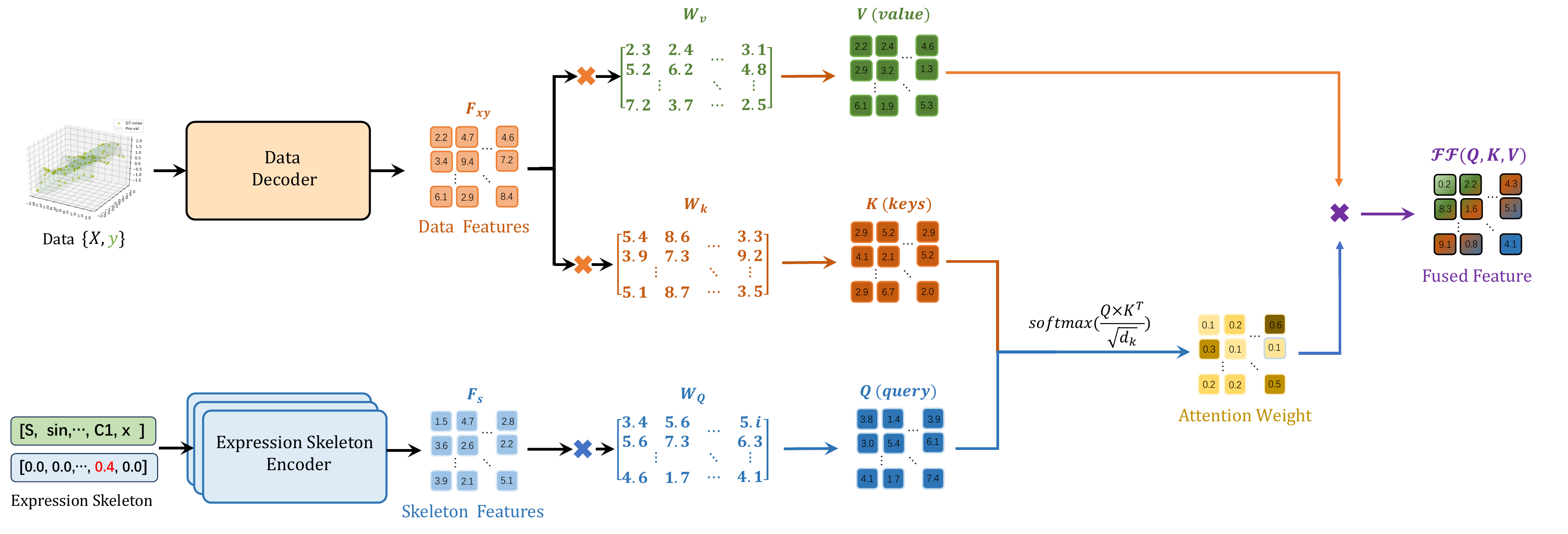}
\caption{
\textcolor{black}{The feature fusion process of MMSR model. The fused features are calculated from the feature information embedded implicitly in the data and expression skeleton.}
}

\label{fig:fusion}
\end{figure*}
\subsubsection{Expression Decoder}

\textcolor{black}{The expression decoder module is primarily responsible for two critical tasks. The first is feature fusion, where aligned data features are combined with skeleton features. The second task is expression generation, which uses these fused features to produce the predicted expression.}

\begin{itemize}
\item \textcolor{black}{\textbf{Feature Fusion} }
\end{itemize}

\textcolor{black}{
Feature fusion is a crucial step in MMSR, as the fused features contain significant information embedded within the data and the expression skeleton. This comprehensive feature enhances the decoder stage, enabling it to generate more accurate and refined target expressions. Specifically, we apply a cross-attention mechanism to achieve this feature fusion, and the fused features are obtained by the following calculation as Eq\ref{qkv}}.
\textcolor{black}{
\begin{equation}
\label{qkv}
\left\{
\begin{aligned}
    & \mathcal{FF}(Q,K,V) &=& \ \text{softmax}\left(\frac{Q \times K^T}{\sqrt{d_k}}\right)V \\
    & \quad \quad  Q &=& F_{s}W_{Q} \\
    & \quad \quad  K &=& F_{xy}W_{K} \\
    & \quad \quad  V &=& F_{xy}W_{V}
\end{aligned}
\right.
\end{equation}
}

\textcolor{black}{Here, $\mathcal{FF}$ denotes the fused features. $F_{xy}$ represents the data feature, while $F_{S}$ refers to the skeleton feature.
$W_{Q}$, $W_{K}$, and $W_{V}$ are mapping matrices responsible for adjusting the dimensions of the corresponding features. $Q$(query) represents the mapped skeleton feature, $K$(keys) can be understood as the description of the data feature, and $V$(value) corresponds to the data feature after matrix mapping. $d_k$ serves as the scaling factor. }

\textcolor{black}{
The detailed process of feature fusion is illustrated in Fig. \ref{fig:fusion}.}

\textcolor{black}{
First, the data $[X,y]$ is passed through the SetTransformer encoder to obtain the data feature $F_{xy}$.This feature $F_{xy}$ is then multiplied by $W_{v}$ and $W_{k}$ to produce the value vector $V$ and the key vector $K$, respectively. While, the skeleton encoder extracts the skeleton feature $F_{s}$, which is subsequently multiplied by $W_{q}$ to generate the query vector $Q$. }

\textcolor{black}{
Next, a dot product operation is performed between the query vector $Q$ and the keys vector $K$. Then, through a Softmax function, the attention weight is obtained.}

\textcolor{black}{
Finally, the attention weight is multiplied by the value vector $V$. So far, we obtained the fused feature that incorporates the data information and the expression skeleton information. }

\begin{itemize}
\item \textcolor{black}{\textbf{Skeleton Generation} }
\end{itemize}

The decoder autoregressively generates the symbols sequence (Skeleton) and the constants sequences given the encoder’s features. We can think of each token as a token in a text generation task. Then, n candidate expressions are obtained by sampling sequentially using beam search according to the predicted probability distribution of the next symbol. And we'll get n constants sequences corresponding to n skeletons of expressions. Next, the constant placeholder `[C-5,..., C5]' in the skeleton is replaced by the corresponding constant to obtain the expression with the rough constant. Finally, we fine-tune the rough constant with BFGS to obtain the optimal expression.

\subsubsection{Modal Alignment Mechanism}
In the multi-modal field,  modal alignment is usually introduced to make the features mapped by the matching modes close, to facilitate the subsequent multi-modal feature fusion, to achieve better results. In order to achieve the modal alignment. Contrastive loss is introduced in mainstream multi-modal models to make the matched data features and skeleton features closer to each other, and the mismatched ones far away from each other in the feature space. Specifically, we align the features of the data obtained by the data point feature extraction extractor with the skeleton features obtained by the skeleton feature extractor. For the data point features, we will get the features of $\mathbb{R}^{b \times k \times h}$ for each batch, where $b$ stands for batch size and $k \times h$ represents the feature size obtained for each data. To facilitate the comparison loss later, we will map the features obtained from each batch from $\mathbb{R}^{b \times k \times h}$ to $\mathbb{R}^{b \times h}$. Similarly, we perform a similar operation for the skeleton feature. Then, the comparison loss is calculated for multiple data and skeleton samples contained in a Batch, and the similarity between positive samples (paired data and skeleton) is expected to be as high as possible, and the similarity between negative samples (unpaired data and skeleton) is expected to be as low as possible. The contrastive loss is formulated as follows.

\begin{equation}
\label{e1}
\begin{aligned}
\mathcal{L}_{CON}= & -\frac{1}{N} \left( \sum_{i}^{N} \log \frac{\exp(x_{i}^{T}y_{j}/\theta)}{\sum_{j=1}^{N}\exp(x_{i}^{T}y_{j}/\theta)} \right. \\
& \left. + \sum_{i}^{N} \log \frac{\exp(y_{i}^{T}x_{j}/\theta)}{\sum_{j=1}^{N}\exp(y_{i}^{T}x_{j}/\theta)} \right)
\end{aligned}
\end{equation}

where $x_i$ and $y_j$ are embeddings of the data in the $i$-th pair and that of the sequence of expression in the $i$-th pair. $N$ is the batch size, and $\theta$ is the temperature to scale the logits.

Futhermore, positive and negative samples are constructed for contrastive learning. Positive samples represent expressions that match with specific data, while negative samples refer to expressions that do not match the given data. The aim of contrastive learning is to enable the model to distinguish between expressions that match (positive samples) and do not match (negative samples) with the data. This recognition capacity is achieved through continual optimization and learning throughout the model's training process. In MMSR, the data encoder is employed to extract features from the data, and the expression skeleton encoder is used to extract features from the expression skeletons, aiming to bring the matching data and expression skeletons as close together as possible in the feature space, whereas ensuring that the mismatches are kept as far apart as possible. By doing so, alignment across different modalities is achieved. This process aids in better learning and understanding of the associations between data features and expression features, enhancing the performance of multimodal solutions to symbolic regression tasks.


\subsection{Joint Training Strategy}
We adopt one-step training strategy to train the network, as shown in Fig. \ref{framework}. The loss function plays a crucial role in directing the training process of deep neural networks, which is imperative for effective deep learning. We design a multi-loss function consisting of three components: cross-entropy loss ($\mathcal{L_{CE}}$), mean squared error(MSE) loss ( $\mathcal{L_{MSE}}$ ), and contrastive loss ( $\mathcal{L_{CON}}$ ). 
\begin{equation}
\label{e2}
\begin{aligned}
\mathcal{L} = \lambda_{1}\mathcal{L_{CE}} +\lambda_{2}\mathcal{L_{MSE}} + \lambda_{3}\mathcal{L_{CON}}
\end{aligned}
\end{equation}
Here, $\mathcal{L_{CE}}$ represents the cross-entropy loss between the predicted expression sequence and the true sequence:
\begin{equation}
\label{e-ce}
\mathcal{L_{CE}}=- \sum_{k}^{M}\log P_{MMSR}(\hat{s}_{k}|\hat{s}_{1:k-1}, X, Y)
\end{equation}
where M is the sequence length and $\hat{s}_{k}$ is the $k$-th predicted symbol, $\hat{s}_{1:k-1}$ denotes the first $k$-1 symbols in the predicted sequence. 

$\mathcal{L_{MSE}}$ represents the MSE loss of the true constants and the predicted constants:
\begin{equation}
\label{e-mse}
\mathcal{L_{MSE}}=\frac{1}{M}\sum_{k}^{M} \frac{(c_k - \hat{c}_k)^2}{(c - \overline{c})^2}
\end{equation} \\
Here, M denotes the length of the predicted constant sequence, $c_k$ denotes the $k$-th true value in the constant sequence, and $\hat{c}_k$ denotes the $k$-th value in the predicted constant sequence. $\overline{c}$ denotes the mean of the real constant sequence. 

$\mathcal{L_{CON}}$ represents the contraction loss over a batch of samples, which is used to make the data features and expression features that match in the feature space closer to each other, and vice versa. The specific formula is given in Eq. \ref{e1}. In MMSR, we achieve modal alignment by introducing contrastive loss $\mathcal{L_{CON}}$ into the loss function.

$\lambda_{1}$, $\lambda_{2}$, $\lambda_{3}$ are hyperparameters that allow us to adjust the proportion of the three losses.

\section{Experiments}
\label{experiments}
\subsection{Dataset}
\begin{itemize}

\item \textcolor{black}{\textbf{Training dataset:}
The magnitude of the training data is 10 million (M), which is generated based on the expression skeleton. After obtaining the expression skeleton by random sampling, the constant encoding is generated and the values [-1,-0.1]$\cup$[0.1,1] are randomly assigned to the locations where the constant placeholders [C-5,...,C5] occurrence in the expression skeleton. Then, we can get an expression $y=f(X)$. For each expression, we collect a random set of data $[X,y]$. Finally, we generated 10M expressions and gained 10M sets of data for MMSR training. For instance, the expression skeleton [sin, $\times$, C1, $x$] is obtained by random sampling and then the constant encoding [0.0, 0.0, 0.23, 0.0] is generated. This yields the encoding of the expression y=sin(2.3$\times$$x$). Then, with a set of $x$-values, a corresponding set of $y$-values can be computed. In this way, we obtained a set of training data. For the sake of a fair comparison, the magnitude of training data for the baseline methods is also set at 10M, which were generated in a similar manner according to the format requirements of each respective method.}


\item \textbf{Testing dataset:} In order to test the performance of MMSR and baselines, we conducted comparative analyses using prevalent datasets, designated as 'Nguyen', 'Keijzer', 'Korns', 'Constant', 'Livermore', 'Vladislavleva', 'R',' Jin', 'Neat',' Others', 'Feynman', 'Strogatz' and 'Black-box'. The three datasets Feynman ', 'Strogatz' and 'Black-box' are from the SRbetch dataset. These datasets contain nearly 400 expressions in total. We believe that they can accurately and comprehensively test the performance of each algorithm.
\end{itemize}

\begin{table*}[htp]
\renewcommand{\arraystretch}{1.1}
\vspace{-0.2cm}
\centering
\caption{The results of performance comparison. At a 0.95 confidence level, a comparison of the coefficient of determination ($R^2$) and the expression complexity(Nodes) conducted between MMSR and four baselines.
\label{tab-r2}}
\resizebox{18.4cm}{!}{
\def\arraystretch{1.2}
    \small
        \bgroup
        \setlength{\tabcolsep}{0.4em}
        \begin{tabular}{c|l|cccccccccc}
\toprule
            \multirow{2}{*}{\bf Group} & \multicolumn{1}{c|}{\multirow{2}{*}{\bf Dataset}} & \multicolumn{2}{c}{\bf MMSR} & \multicolumn{2}{c}{\bf TPSR\cite{shojaee2024transformer}} & \multicolumn{2}{c}{\bf End2End\cite{kamienny2022end}} & \multicolumn{2}{c}{\bf NeSymReS\cite{DBLP:conf/icml/BiggioBNLP21}} & \multicolumn{2}{c}{\bf SymbolicGPT\cite{DBLP:journals/corr/abs-2106-14131}}  \\
            \cline{3-12}
& & $R^2$ &Nodes  & $R^2$ &Nodes& $R^2$ &Nodes& $R^2$ &Nodes & $R^2$ &Nodes  \\ 
\cmidrule(lr){2-2}
\cmidrule(lr){3-4}
\cmidrule(lr){5-6}
\cmidrule(lr){7-8}
\cmidrule(lr){9-10}
\cmidrule(lr){11-12}

\multirow{10}{*}{\rotatebox{90}{Standards}}

& Nguyen    & $\textbf{0.9999}_{\pm0.001}$&\textbf{14.5} & $0.9948_{\pm0.002}$ &16.0& $0.8814_{\pm0.004}$&16.3& $0.8568_{\pm0.003}$&18.2& $0.6713_{\pm0.005}$ &21.6 \\
& Keijzer   & $\textbf{0.9983}_{\pm0.003}$ &\textbf{16.3}& $0.9828_{\pm0.003}$&20.6& $0.8134_{\pm0.005}$&18.4 & $0.7992_{\pm0.003}$&21.3 & $0.6031_{\pm0.004}$  &24.5\\
& Korns     & $\textbf{0.9982}_{\pm0.003}$ &\textbf{19.2} &$ 0.9325_{\pm0.004}$&22.9& $0.8715_{\pm0.004}$&23.4  & $0.8011_{\pm0.005}$ &24.1& $0.6613_{\pm0.005}$  &29.2\\
& Constant  & $\textbf{0.9986}_{\pm0.002}$&\textbf{24.5}  & $0.9319_{\pm0.002}$ &35.3& $0.8015_{\pm0.003}$ &28.3 & $0.8344_{\pm0.003}$  &32.9& $0.7024_{\pm0.004}$  &38.5 \\
& Livermore  & $\textbf{0.9844}_{\pm0.003}$&\textbf{29.4} & $0.8820_{\pm0.004}$&38.2& $0.7015_{\pm0.004}$&32.2& $0.6836_{\pm0.005}$&36.2 & $0.5631_{\pm0.0005}$ &41.2\\
& Vladislavleva  & $\textbf{0.9862}_{\pm0.003}$&\textbf{21.7} & $0.9028_{\pm0.005}$&24.6 & $0.7422_{\pm0.005}$&22.2 & $0.6892_{\pm0.004}$ &27.3& $0.5413_{\pm0.004}$  &36.6\\
& R  & $\textbf{0.9924}_{\pm0.004}$&16.4 & $0.9422_{\pm0.003}$&\textbf{16.2} & $0.8512_{\pm0.004}$ &19.5& $0.7703_{\pm0.005}$&19.9 & $0.7042_{\pm0.005}$ &25.2 \\
& Jin  & $\textbf{0.9943}_{\pm0.003}$ &\textbf{28.3}& $0.9826_{\pm0.004}$&29.5 & $0.8611_{\pm0.004}$&29.8 & $0.8327_{\pm0.003}$&32.2& $0.7724_{\pm0.006}$&36.9\\
& Neat  & $\textbf{0.9972}_{\pm0.004}$&17.3& $ 0.9319_{\pm0.002}$ &\textbf{16.4}& $ 0.8044_{\pm0.004}$&19.7 & $ 0.7596_{\pm0.005}$&20.6& $ 0.6377_{\pm0.005}$ &26.4\\
& Others  & $\textbf{0.9988}_{\pm0.002}$ &20.6& $0.9667_{\pm0.002}$&22.5& $0.8415_{\pm0.003}$&22.3 & $0.8026_{\pm0.003}$&23.5 & $0.7031_{\pm0.004}$&31.8 \\
\midrule
\multirow{3}{*}{\rotatebox{90}{SRBench}}
& Feynman  & $\textbf{0.9913}_{\pm0.002}$&\textbf{20.8 }&$0.8928_{\pm0.004}$ &21.3&$0.7353_{\pm0.004}$&22.0  & $0.7025_{\pm0.005}$&22.4 &$0.5377_{\pm0.005}$&26.8\\
& Strogatz  & $\textbf{0.9819}_{\pm0.003}$&\textbf{21.6} &$0.8249_{\pm0.002}$&24.4 &$0.6626_{\pm0.003}$&25.4  & $0.6022_{\pm0.003}$&28.1 &$0.5229_{\pm0.004}$&32.6\\
& Black-box & $\textbf{0.9937}_{\pm0.004}$&\textbf{26.7} &$0.8753_{\pm0.004}$ &29.3&$0.6925_{\pm0.004}$ &31.2 & $0.6525_{\pm0.005}$&33.9 &$0.5833_{\pm0.005}$&37.4\\
\cline{2-7} 
\toprule
\multirow{1}{*}{\rotatebox{90}{ }}
 & Average & $\textbf{0.9934}$&\textbf{21.3} &$0.9264$ &24.4&$0.7892$ &23.9 & $0.7528$ &26.2& $0.6311$&32.4\\
 \toprule
\end{tabular}
\egroup
}
\end{table*} 
\subsection{Compared Methods}
In order to test the performance of our algorithm against similar algorithms, we selected some representative "pre-training-based" symbolic regression algorithms in recent years. 
\begin{itemize}

    \item \textbf{TPSR}\cite{shojaee2024transformer}: One uses a pre-trained model as a policy network to guide the MCTS process, which greatly improves the search efficiency of MCTS.
   \item \textbf{End2End}\cite{kamienny2022end}: End2End further encodes the constants so that the model generates the constants directly when generating the expression, abandoning the constant placeholder `C' used by the previous two algorithms.  
    \item \textbf{NeSymReS}\cite{DBLP:conf/icml/BiggioBNLP21}: Based on SymbolicGPT, the algorithm treated each `operator' (e.g. `sin', `cos') as a token. The sequence of expressions is generated in turn. The setup is more reasonable.   
    \item \textbf{SymbolicGPT}\cite{DBLP:journals/corr/abs-2106-14131}: The classical algorithm for pre-trained models, which treats the symbolic regression problem as a translation problem. Treat a `single letter' (e.g. `s', `i', `n') as a token. Generating expressions sequentially.
\end{itemize}

\subsection{Evaluation Indicators}
\subsubsection{Fitting Ability}

Coefficient of determination ($R^2$) is utilized as the metric to evaluate the performance. Formally, $R^2$ is calculated as follows:
\begin{equation}
\label{e3}
\begin{aligned}
R^2 = 1-\sum_{i=1}^n \frac{(y_i - \hat{y_i})^2}{(y_i-\overline{y})^2}
\end{aligned}
\end{equation}
Where $n$ is the number of sample points, $y_i$ is the $i^{th}$ true value, $\hat{y_i}$ is the $i^{th}$ predicted value, and $\overline{y}$ is the mean of the true y values.

\subsubsection{Expression complexity}

Symbolic regression aims to find mathematical expressions that are concise and explanatory. Therefore, when evaluating the performance of different symbolic regression algorithms, the complexity of the expressions generated by the algorithms always be considered. 

Therefore, in this work, we evaluate the complexity of each method by counting the length of the preorder traversal of the final expression produced by each method, i.e., the number of symbols. For example in Fig. \ref{fig11} expression $y = 4.2 * x_2 + \ sin (x_1) $, the first sequence traversal of expression binary tree  is [+,$*$, c1, $x_2$, sin, $x_1$], a total of 6 nodes (symbols), so the expression of complexity is 6.
\subsubsection{Model Complexity}
We measure the complexity of a model in terms of the number of model parameters and time complexity (inference time).
\begin{itemize}
    \item \textbf{Number of parameters}. The number of all parameters in the model that need to be optimized updated.
    \item \textbf{Time complexity}. The amount of time, in seconds, from the input to the final result.    
\end{itemize}

\begin{figure}[htp]
\centering
\includegraphics[width=88mm]{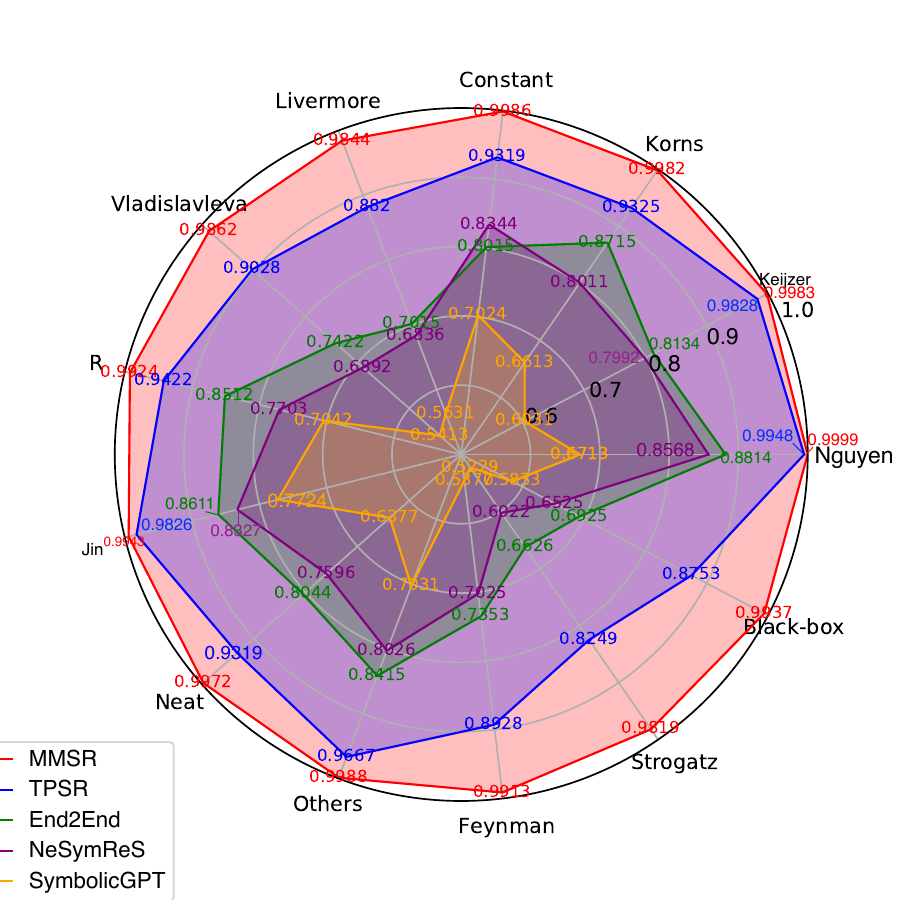}
\caption{
Radar charts demonstrating the fitting ability ({$R^2$}). It  can be found that MMSR achieved the best fitting performance among all the comparison methods.
}
\label{fig1}
\end{figure}
\subsection{Results}

\subsubsection{Performance Evaluation of MMSR}

In Regression Symbolic, $R^2$ is the most basic and important metric to reflect the fitting ability of the algorithm.
In the paper, each expression in the test set is tested 20 times, and then the average of its $R^2$ is taken as the final result, with a confidence interval of 0.95. The performance comparison between our algorithm and four baselines is shown in Table \ref{tab-r2} and visualised with radar chart in Fig.\ref{fig1}  intuitively. From the experiment results, we can find that MMSR achieved the best fitting performance among all the comparison methods.

Furthermore, we run 20 times for each test sample of all datasets and subsequently compute the average complexity of the resulting expressions. Detailed statistical results regarding the average complexity of each dataset are listed in Table \ref{tab-r2}. The expressions obtained through the proposed MMSR method have the lowest average complexity. This makes the expressions extracted from the data using the MMSR method easier to interpret compared to other methods. This is the goal that MMSR is dedicated to, obtaining more concise and more accurate expressions.
\begin{figure*}[htpb]
    \centering
    \setlength{\belowcaptionskip}{-0.3cm} 
      \subfloat[]{
      \includegraphics[width=0.45\linewidth]{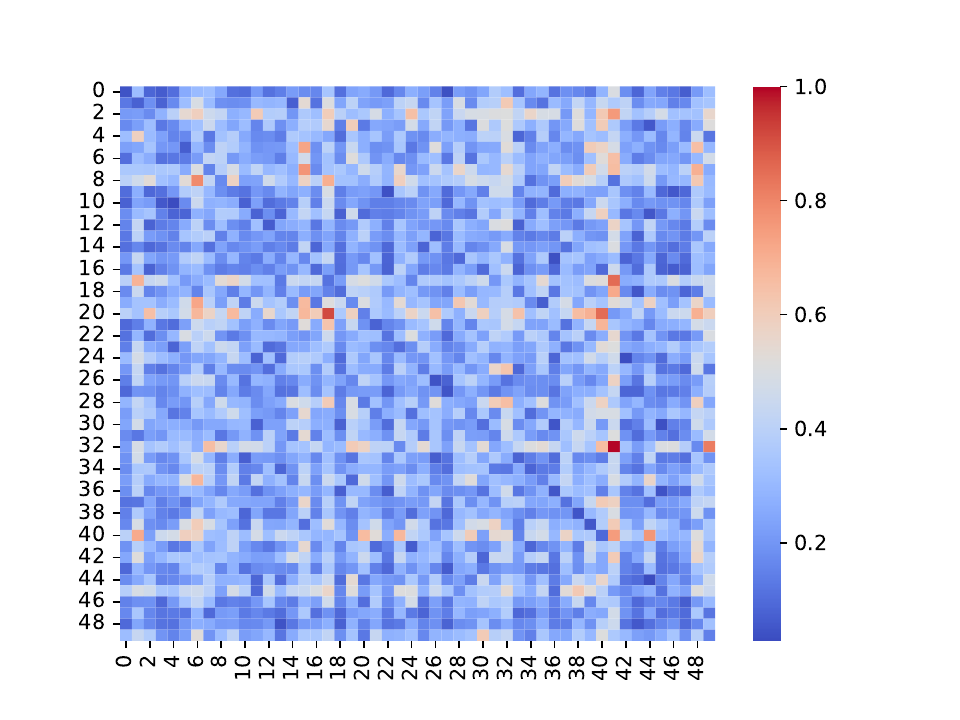} \label{fig3a}}  
        \subfloat[]{
        \includegraphics[width=0.45\linewidth]{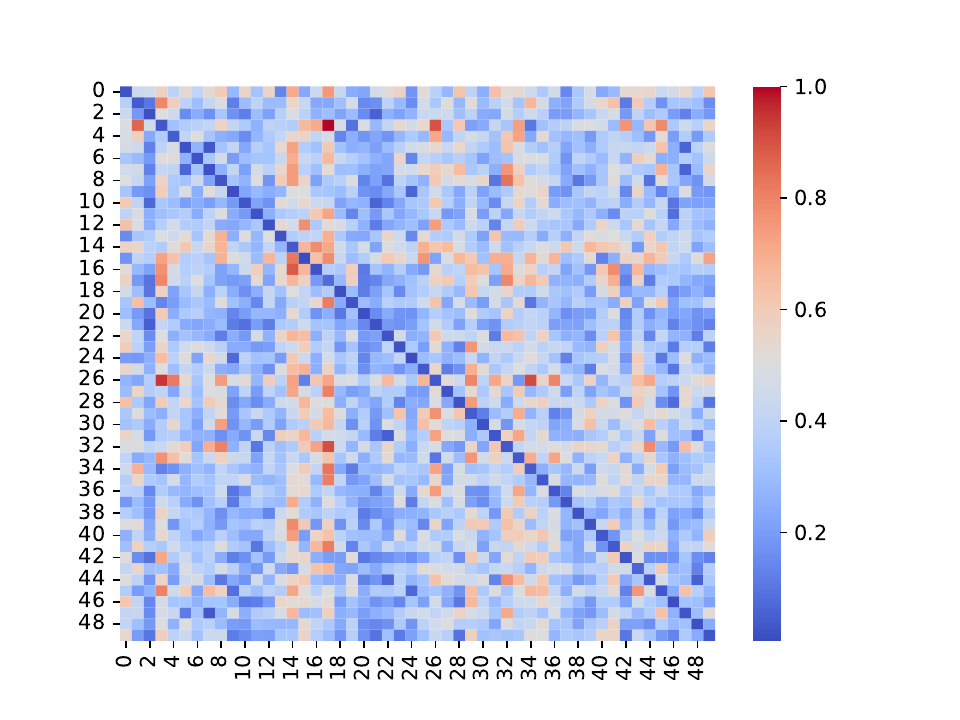}\label{fig3b}}   
	  \caption{Distance of data feature pairs after normalization. (a) without modal alignment, (b) with modal alignment.
As we can see from the figure, under the premise that other configurations are completely the same, before the introduction of contrastive learning (a), the cosine distance between the matched data features and the skeleton features extracted by the model in the feature space is not significantly smaller than that of the unmatched ones (the values on the diagonal are not significantly smaller than the others). However, after introducing contrastive learning (b), we can clearly see that the values on the diagonal are significantly smaller than the values at other positions. That's what we expected.}
\label{fig3} 
\end{figure*}
\subsubsection{Effect of Modal Alignment on Features}
To verify whether the extracted features before and after the introduction of contrastive learning are as expected: Data and skeleton features that match each other are close to each other in the feature space, whereas those that do not match are far away from each other.
We first randomly generate 50 expression samples, feed these samples into the model without contrastive learning, and then obtain 50 pairs of features for the data and skeleton. We calculate the cosine distance between the extracted 50 data features and 50 skeleton features and then obtain the 50*50 cosine distance matrix A. Finally, we normalize the matrix A and visualize it to get Fig. \ref{fig3a}. Similarly, we feed these samples into a model that uses contrastive learning, resulting in heatmap \ref{fig3b}.

From Fig. \ref{fig3}, we can see that there is indeed a large change in the extracted features before and after the introduction of contrastive learning. In Fig. \ref{fig3a}, we can see that the data on the diagonal is not significantly smaller than the other positions, indicating that the matching data features and skeleton features are not the closest to each other in the feature space. On the contrary, we can clearly find in Figure 3b that the data on the diagonal is significantly smaller than the other positions, which indicates that our contrastive learning plays a role. The features of the two modalities, data and skeleton, have been well aligned.

\subsubsection{Ablation Study of Modal Alignment}
In MMSR, we achieve modal alignment by introducing contrastive loss ($\mathcal{L_{CON}}$) into the loss function  ($\mathcal{L}$), and denote the degree of the contribution of modal alignment by $\lambda_3$. To investigate the effectiveness of modal alignment to the symbolic regression tasks, we conduct an ablation experiment consisting of three groups. In each group, the settings of the parameters $\theta$, $\lambda_1$ and $\lambda_2$ were kept consistent, where $\theta$ = 0.07, $\lambda_1$ = $\lambda_2$ =1. And the parameter $\lambda_3$ was set to 0.0, 0.1 and 1.0 respectively. When $\lambda_3$ = 0.0, it indicates that modal alignment is not used. When $\lambda_3$ = 0.1 or 1.0, it indicates that the modal alignment is applied, and the larger the value of $\lambda_3$, the greater the role that modal alignment is plays in the MMSR method.

\begin{table}[ht]
\center
\caption{The results of ablation experiments examining the effect of modal alignment.}
\begin{tabular}{cccc}
\toprule
Dataset& \multicolumn{3}{c}{$R^2$}\\ \toprule
\multirow{4}{*}{Hyperparameters}

&$\theta=0.07$ & $\theta=0.07$ &$\theta=0.07$ \\ 
&$\lambda_1=1.0$ & $\lambda_1=1.0$ & $\lambda_1=1.0$ \\ 
&$\lambda_2=1.0$ & $\lambda_2=1.0$ & $\lambda_2=1.0$ \\ 
&$\lambda_3=1.0$ & $\lambda_3=0.1$ & $\lambda_3=0.0$ \\ 
      \cmidrule(lr){1-4}
Nguyen        & 0.9999 & 0.9994 & 0.9701 \\
Keijzer       & 0.9983 & 0.9724 & 0.9605 \\
Korns         & 0.9982 & 0.9842 & 0.9364 \\
Constant      & 0.9986 & 0.9024 & 0.8883 \\
Livermore     & 0.9844 & 0.9388 & 0.9034 \\
Vladislavleva & 0.9862 & 0.9603 & 0.9242 \\
R             & 0.9924 & 0.9831 & 0.9683 \\
Jin           & 0.9943 & 0.9711  & 0.9422 \\
Neat          & 0.9972 & 0.9633 & 0.9248 \\
Others        & 0.9988 & 0.9555 & 0.9146 \\
Feynman       & 0.9913 & 0.9735  & 0.9644 \\
Strogatz      & 0.9819 & 0.9533 & 0.9272 \\ 
Black-box     & 0.9937 & 0.9655 & 0.9028 \\ 
 \cline{1-4}
Average & \textbf{0.9934} & \textbf{0.9633} & \textbf{0.9328}\\
\toprule
\label{tab:1}
\end{tabular}
\end{table}
\begin{figure*}[htp]
    \centering
      \subfloat[]{
      \includegraphics[width=0.99\linewidth]{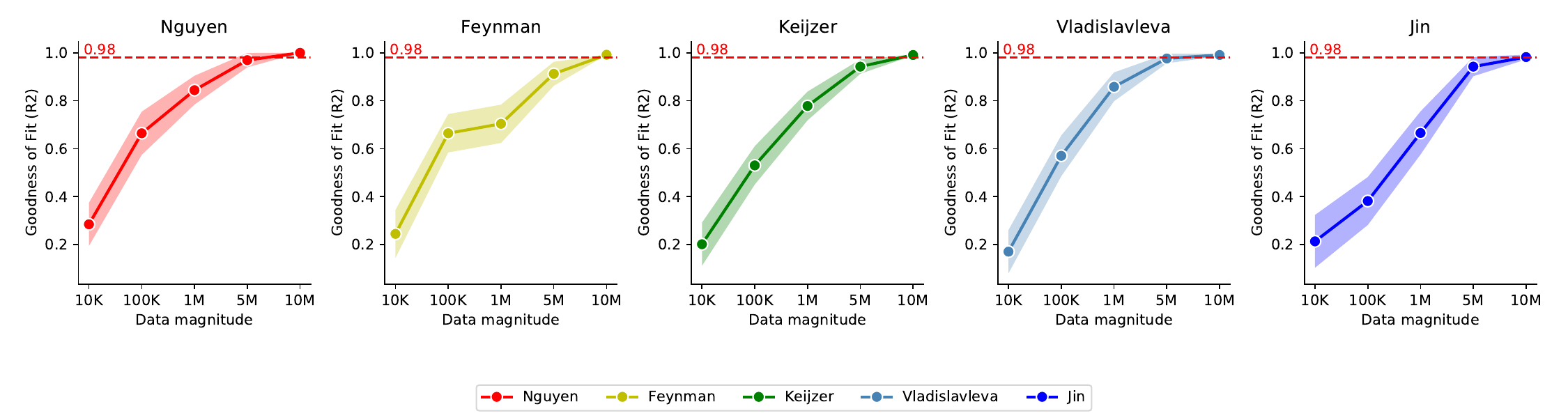} \label{fig4a}}\\
        \subfloat[]{
        \includegraphics[width=0.99\linewidth]{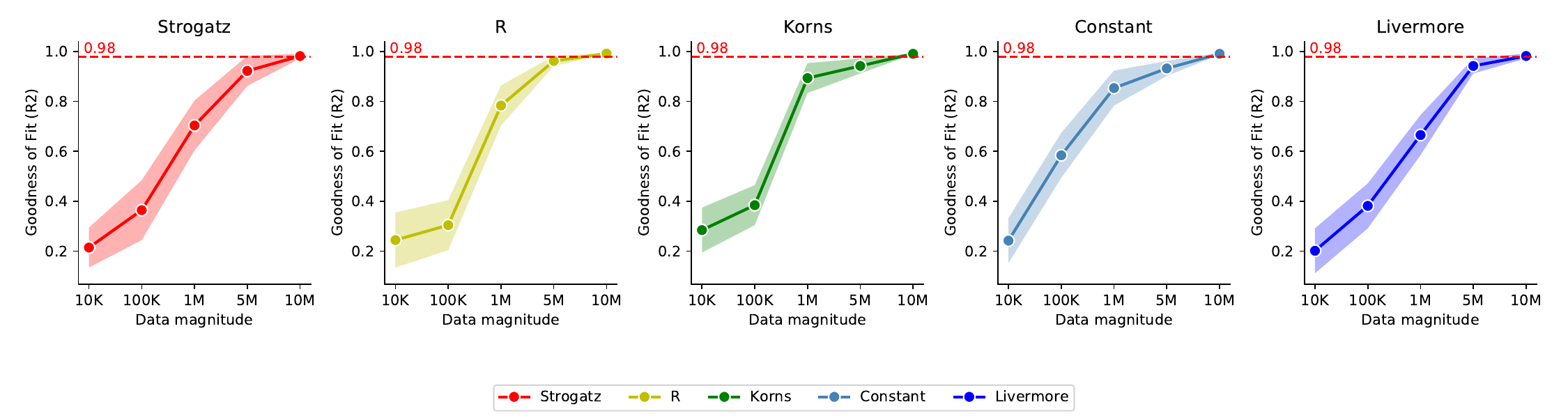}\label{fig4b}}\\
        \subfloat[]{
      \includegraphics[width=0.80\linewidth]{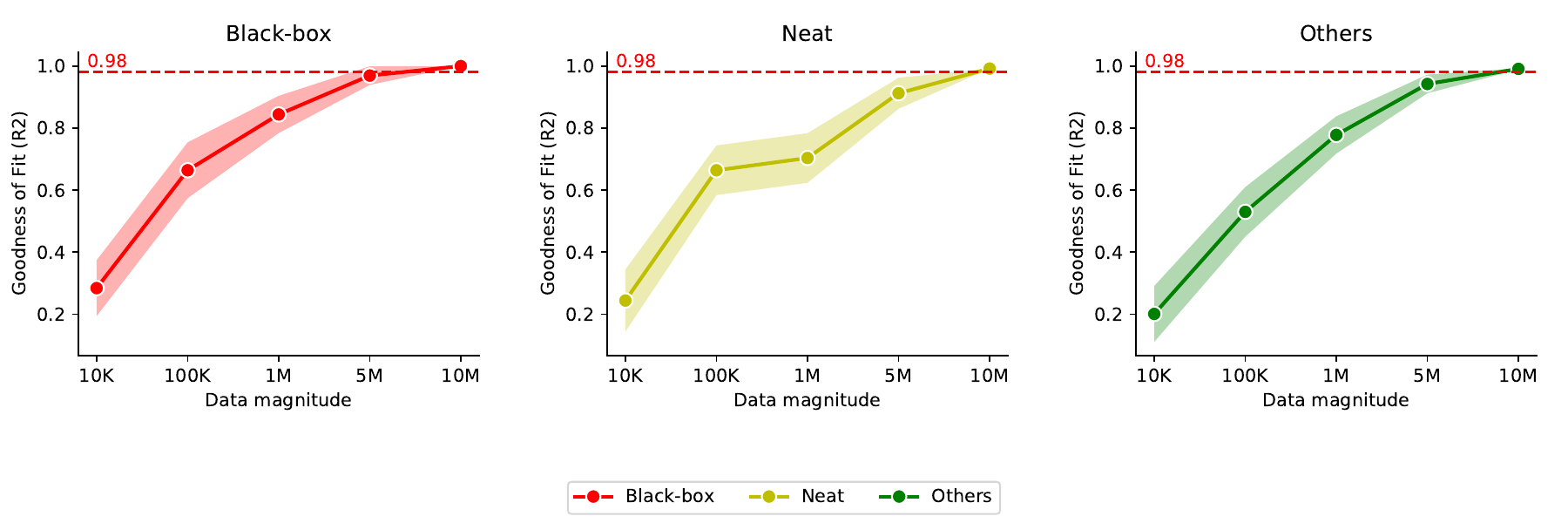} \label{fig4c}}
	  \caption{
  Effect of dataset size on fitting performance. From the above plots, It can be found that as the size of the training data increases, the performance of the model on each dataset continues to improve. When the data size reaches 10M, the model has achieved satisfactory results on the selected data sets, and the average $R^2$ on each data set is greater than 0.98. }
\label{fig4} 
\end{figure*}
The specific experimental results are shown in Table \ref{tab:1}. Under the same conditions of parameters $\theta$, $\lambda_1$ and $\lambda_2$, the best overall performance is achieved at $\lambda_3$ = 1.0, which is much better than the performance at $\lambda_3$ = 0.0. It is not difficult to find that, as the weight $\lambda_3$ of contrastive loss ($\mathcal{L_{CON}}$) decreases, the performance ($R^2$) of MMSR also decreases gradually. This experimental result fully illustrates that modal alignment positively affects the performance of the MMSR model.
This beneficial effect arises from the fact that the contrastive loss can bring the matched data features and the skeleton features closer together in the feature space, and more importantly, it keeps the non-matched data features and the skeleton features farther away from each other. This results in better integration of the latter two features.

In summary, from the above ablation experiments, we can conclude that modality alignment plays a positive role for symbolic regression tasks.
\begin{figure*}[ht]
\centering
\includegraphics[width=1.0\linewidth]{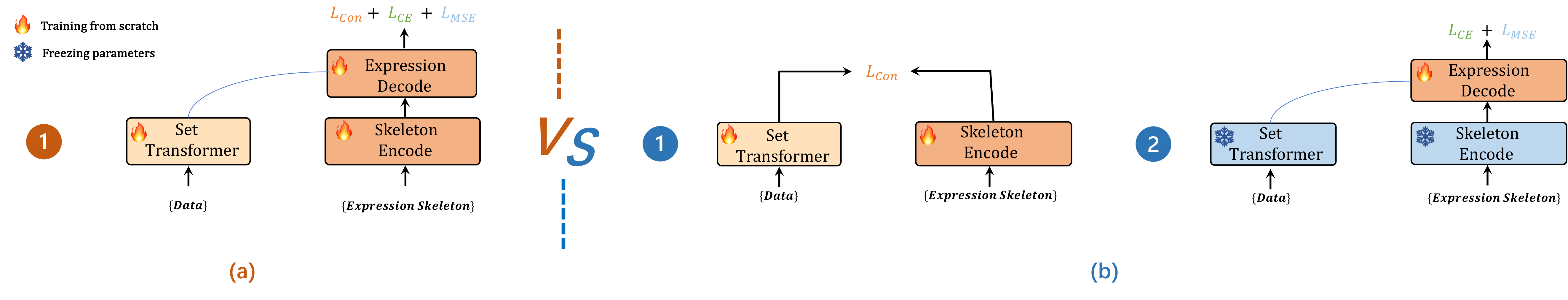}
\caption{
(a) represents joint training. (b) represents separate training. Comparison plots of the two contrastive learning introduction strategies. From the diagram above, we can intuitively see the difference between the Joint training strategy and the Separate training strategy. The Joint training strategy trains the three modules together, while the Separate training strategy trains the two encoders and then the decoder.
}
\label{fig6}
\end{figure*}
\begin{table}[ht]
\center
\caption{The results of ablation experiments examining the effect of additional constant encoding.}
\begin{tabular}{cccc}
\toprule
Dataset& \multicolumn{3}{c}{$R^2$}\\ \toprule
\multirow{4}{*}{Hyperparameters}
& $\theta=0.07$ & $\theta=0.07$ & $\theta=0.07$ \\ 
& $\lambda_1=1.0$ & $\lambda_1=1.0$ & $\lambda_1=1.0$ \\ 
& $\lambda_2=1.0$ & $\lambda_2=0.1$ & $\lambda_2=0.0$ \\ 
& $\lambda_3=1.0$ & $\lambda_3=1.0$ & $\lambda_3=1.0$ \\ 

\cmidrule(lr){1-4}
Nguyen        & 0.9999 & 0.9936 & 0.9226 \\
Keijzer       & 0.9983 & 0.9435 & 0.8935 \\
Korns         & 0.9982 & 0.9437 & 0.9025 \\
Constant      & 0.9986 & 0.9753 & 0.9382 \\
Livermore     & 0.9844 & 0.9346 & 0.8633 \\
Vladislavleva & 0.9862 & 0.9373 & 0.8824 \\
R             & 0.9924 & 0.9836 & 0.9146 \\
Jin           & 0.9943 & 0.9917 & 0.9538 \\
Neat          & 0.9972 & 0.9626 & 0.9248 \\
Others        & 0.9988 & 0.9935 & 0.9736 \\
Feynman       & 0.9913 & 0.9625 & 0.9027 \\
Strogatz      & 0.9819 & 0.9264 & 0.8552 \\
Black-box     & 0.9937 & 0.9482 & 0.8636 \\
\cline{1-4}
Average & \textbf{0.9934} & \textbf{0.9613} & \textbf{0.9069} \\
\toprule
\label{tab:lmse}
\end{tabular}
\end{table}
\subsubsection{Ablation Study of Constant Encoding}
To assess the impact of incorporating additional constant encoding on the performance of the algorithm, we conducted ablation experiments on the $\mathcal{L_{MSE}}$ component of the loss function $\mathcal{L}$, where $\lambda_2$=0 represents the absence of constant encoding. The ablation experiments also consisted of three groups. In each group, the settings of the parameters $\theta$, $\lambda_1$ and $\lambda_3$ were kept consistent, where $\theta$ = 0.07, $\lambda_1$ = $\lambda_3$ =1. And the parameter $\lambda_2$ was set to 0.0, 0.1 and 1.0 respectively.
The experimental results can be found in Table \ref{tab:lmse}. The results indicate that adding constant encoding significantly enhances the performance of the MMSR algorithm. By assigning a unique label to each data set, training becomes more straightforward. For instance, in NeSymReS, the sequence [sin, *, C, x] could represent functions such as sin(4x), sin(4.3x), etc. Despite these functions having a similar structural appearance, they can exhibit considerable differences. Introduction of constant encoding facilitates a one-to-one mapping between data instances and their corresponding expressions, effectively reducing the ambiguity of multiple data sets correlating to the same symbolic sequence, thereby aiding in network training. Moreover, dedicating a unique encoding to constants addresses issues where expressions might share high token-level similarities but differ significantly in achieving specific equation-driven goals like fitting accuracy. This issue typically arises when relying solely on cross-entropy loss between the generated and actual sequences.

\subsubsection{Effect of the Size of the Training Data}
To test the impact of pre-training data size on model test performance, we train our MMSR model on increasingly larger datasets. More specifically, we use datasets consisting of 10K, 100K, 1M, 5M, and 10M samples to train a model respectively. Except for the amount of data, every other aspect of training is the same as described above, and all are trained for the same epochs. 
From the trend in Fig. \ref{fig4}, we can see that the performance of the model has been greatly improved as the data set continues to grow. We can see that when the data size reaches 5M, the average $R^2$ has reached more than 0.9, and when the data size reaches 10M, The model achieves an average $R^2$ of over 0.98 on each dataset. Although the growth rate of average $R^2$ is relatively slow with the increase of data, it is still increasing. Therefore, it can be inferred that as we continue to increase the size of the training data, the performance of the model will still reach higher levels.
\subsection{Effect of the Introduction Strategy of Contrastive Learning}
It is very important to use contrastive learning for modality alignment in multi-modality. During training, there are two strategies for introducing contrastive learning: the joint training strategy and the separate training strategy. A comparison of these two strategies is shown in Fig.\ref{fig6}.
\begin{itemize}
    \item \textbf{Joint training strategy}: The contrast loss is used as a part of the total loss, and the feature extraction module and the feature fusion module are trained at the same time.
    \item \textbf{Separate training strategy}: First, train a feature extraction module with contrastive learning, freeze it, and then train a feature fusion module separately. 
\end{itemize}
\begin{figure*}[htp]
    \centering
      \subfloat[]{
      \includegraphics[width=0.99\linewidth]{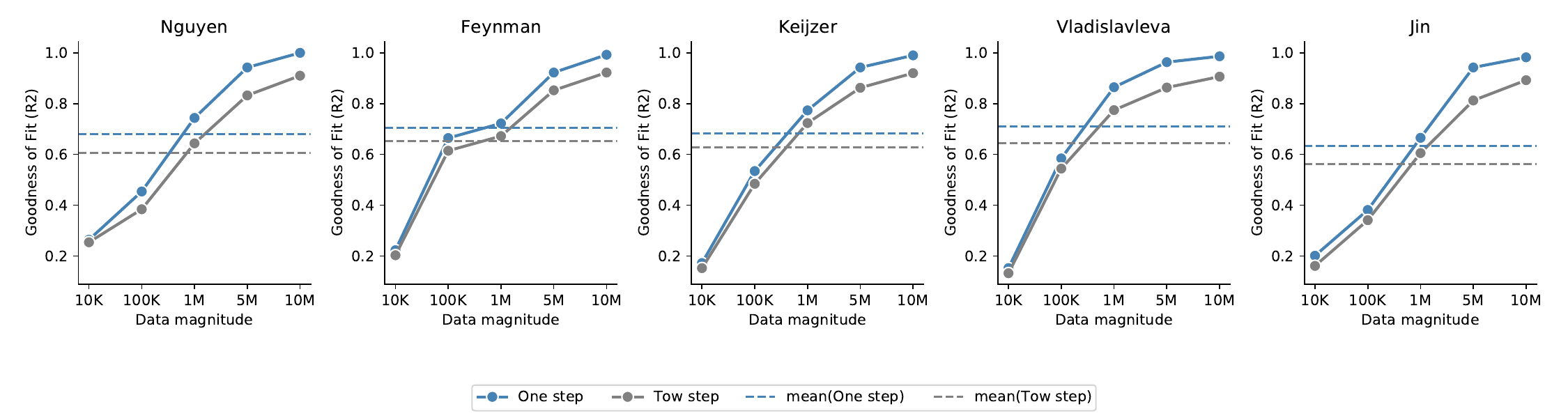} \label{fig5a}}\\
        \subfloat[]{
        \includegraphics[width=0.99\linewidth]{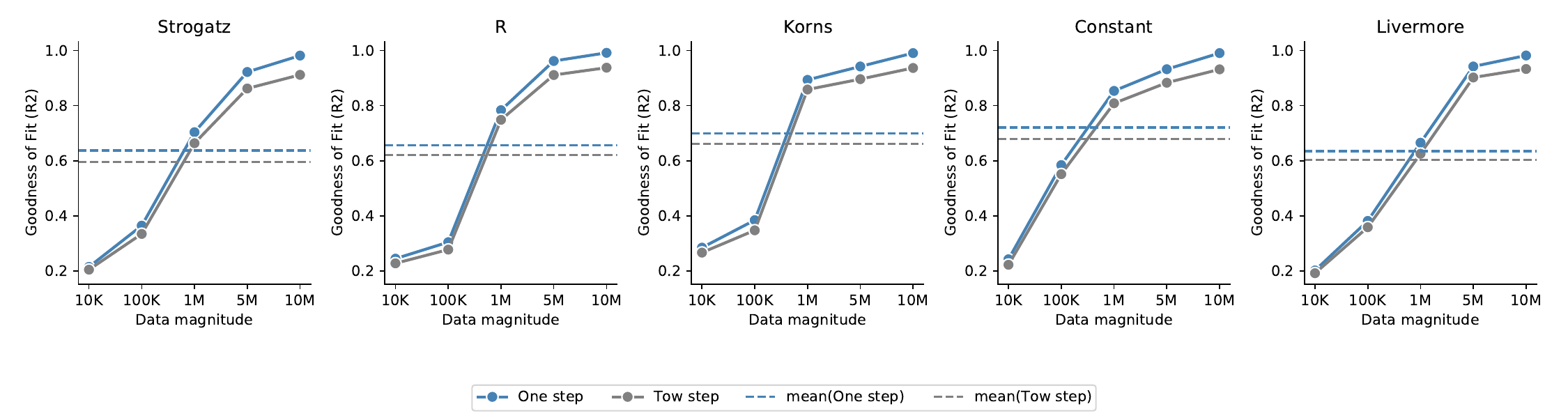}\label{fig5b}}\\
        \subfloat[]{
      \includegraphics[width=0.77\linewidth]{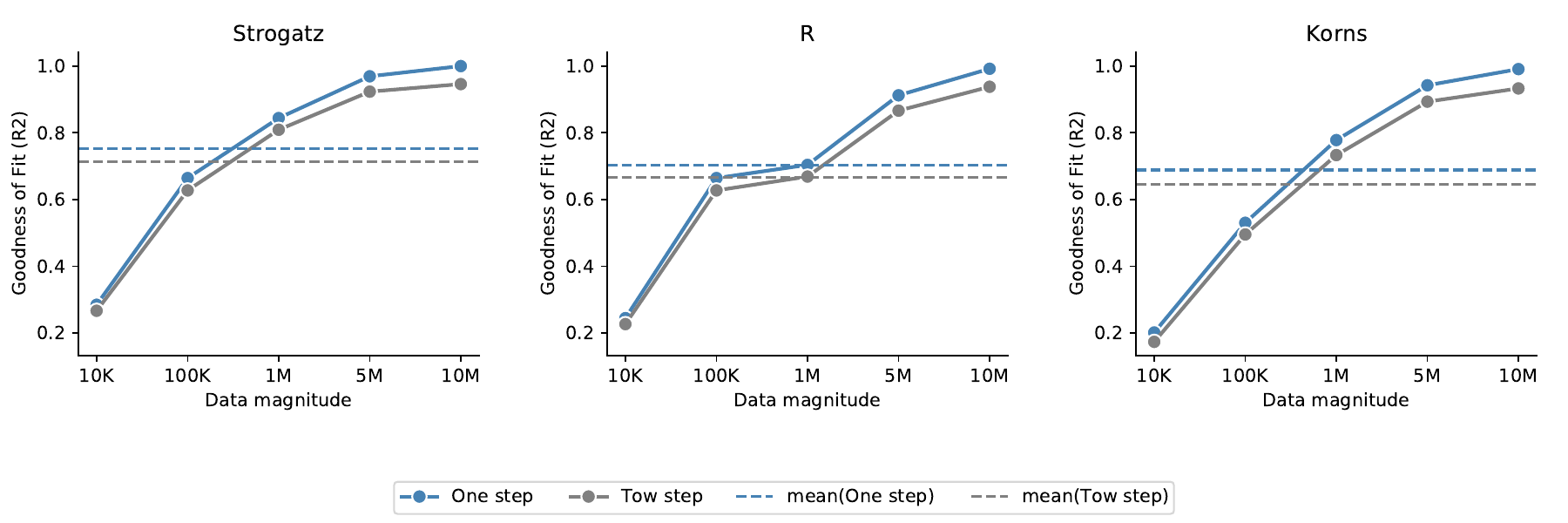} \label{fig5c}}
	  \caption{
  Performance comparison of joint training and separate training with different training set sizes. From the above plots, it can be seen that the algorithm's performance increases as the training data size increases. However, under the same data size, the model using the joint training strategy is better than the model using the separate training strategy. And the larger the amount of data, the more obvious this trend. }
\label{fig5} 
\end{figure*}
 
 To prove the superiority of the proposed joint training strategy, we conduct comparative experiments on all datasets. From Fig.\ref{fig5}, it can be found that the joint training strategy is better than the separate training strategy in terms of effect. As the size of the dataset increases, this advantage becomes more obvious. 

We posit that the underlying reason for this phenomenon lies in the implementation of the joint training strategy, wherein contrastive loss ( $\mathcal{L_{CON}}$ )
is simultaneously optimized alongside cross-entropy loss ($\mathcal{L_{CE}}$) and MSE loss ( $\mathcal{L_{MSE}}$ ). During the training process, the feature extraction module and the feature fusion module are inherently interdependent. This interrelation necessitates that the feature extraction module not only brings the extracted data and skeleton features closer together in the feature space but also ensures that the extracted features are optimally aligned with the requirements of the subsequent feature fusion module. Consequently, this synergy enables a reduction in overall loss. In summary, the joint training strategy facilitates a more effective interaction between each module, enhancing their collaborative performance during concurrent training.

\subsection{Model Complexity Evaluation}
We measure the complexity of the algorithm in terms of the number of model parameters and the time complexity. In this paper, the transformer of all models uses 16 attention heads and an embedding dimension of 512, and the number of decoder layers is 16. (Note that for MMSR, the first four layers are skeleton encoders. No cross-attention is employed. The last four layers are expression decoders). Each method differs in the data encoder, but there is little difference in the overall number of parameters of the model, and the specific parameter scale is shown in Table \ref{tab-complexity}(Number Size of Parameters). 
We counted the average inference time for each algorithm on the Nguyen dataset. We test each expression 20 times, record the time it takes to infer it, and then average it overall. The specific results are shown in Table \ref{tab-complexity}(Average Time of inference.

From the comparison of model complexity, it is observable that, aside from the TPSR method having the longest average running time, the proposed MMSR achieves a higher $R^2$ compared to other methods under the same conditions of parameter quantity and execution time, demonstrating superior performance.

\begin{table}[htpb]
\center
\vspace{-0.2cm}
\caption{Complexity evaluation of MMSR and the state-of-the-art methods.}
\label{tab-complexity}
\resizebox{9.0cm}{!}{
\begin{tabular}{lcccccc}
 \toprule
 \multicolumn{1}{c}{Methods} & \multicolumn{1}{c}{Number Size} & \multicolumn{1}{c}{Average Time} & \multicolumn{1}{c}{Average $R^2$} \\
 \multicolumn{1}{c}{} & \multicolumn{1}{c}{of Parameters} & \multicolumn{1}{c}{of inference} & \multicolumn{1}{c}{of test dataset} \\
 \toprule
 MMSR & 86M & 26.6s & \textbf{0.9999} \\
 TPSR\cite{shojaee2024transformer} & 86M & 86.3s & 0.9948 \\
 End2End\cite{kamienny2022end} & 86M & \textbf{24.4s} & 0.8814 \\
 NeSymReS\cite{DBLP:conf/icml/BiggioBNLP21} & \textbf{84M} & 27.3s & 0.8568 \\
 SymbolicGPT\cite{DBLP:journals/corr/abs-2106-14131} & 88M & 28.4s & 0.6713 \\
 \toprule 
\end{tabular}
}
\end{table}

\section{Discussion and Conclusion}
In this paper, we propose a symbolic regression algorithm, MMSR. We solve the SR problem as a purely multi-modal problem and obtain decent experimental results. This also proves that it is practical to treat the SR problem as a multi-modal task. Specifically, we treat the data and expression skeleton as two different modalities (e.g. image and text). For the data part, we use SetTransformer for the feature extractor, and for the expression skeleton part, we use the first few layers of the transformer as the feature extractor. Specifically, we introduce contrastive learning for modal feature alignment for better feature fusion. The ablation experiments on more than a dozen datasets show that the introduction of contrastive learning has a great improvement in the performance of the algorithm. In addition, we adopt the One-step scheme(the contrastive loss is trained along with other losses during training) in MMSR. Our experiments show that the One-step scheme is more competitive than the Tow-step scheme, which first trains a feature extraction module with contrastive loss and then trains the feature fusion module separately. In MMSR, alleviating expressions may exhibit high token-level similarities but are suboptimal with respect to equation-specific objectives such as fitting accuracy. Our decoder not only generates the sequence of symbols that make up the skeleton of an expression but also directly generates a set of constants. This set of constants is the same length as the symbol sequence, we take the constant with the same index as the placeholder 'C' as the value at 'C', and then the constants are further refined with BFGS to obtain the final expression.

Our algorithm is expected to open new application prospects of multi-modal technology. It promotes more multi-modal correlation techniques to be applied to the field of symbolic regression. We can even use symbolic regression techniques to re-feed multi-modal domains. 

At present, although MMSR has achieved good results, it also has many imperfect. For example, its anti-noise performance is poor. The type of symbol used cannot be changed once the training is complete. The maximum number of support variables must also be specified in advance and cannot be expanded once the training is complete.

Next, we will continue to explore the deep fusion of multi-modal techniques and symbolic regression techniques. It also attempts to solve some of the existing problems of MMSR mentioned above.
\label{conclusion}

\section*{Acknowledgements}
This work was supported by the National Natural Science Foundation of China under Grant 92370117, and CAS Project for Young Scientists in Basic Research under Grant YSBR-090.








\bibliographystyle{elsarticle-num} 
\bibliography{elsarticle}

\newpage
\onecolumn
\appendix
\renewcommand\thesection{\Alph{section}} 
\counterwithin{figure}{section} 
\counterwithin{table}{section}

\section{Appendix: Detailed Settings of hyperparameters during training and inference}

\begin{table*}[htbp]
\centering
{
\caption{Hyperparameters of SetTransformer}
\begin{tabular}{lc}
\textbf{hyperparameters} & \textbf{Numerical value}\\ 
\toprule
 \textbf{N\_p}  & 0\\
 \textbf{activation}  & `relu'\\
 \textbf{bit16}  & True\\
 \textbf{dec\_layers}  & 5\\
 \textbf{dec\_pf\_dim}  & 512\\
 \textbf{dim\_hidden}  & 512\\
 \textbf{dim\_input}  & 3\\
 \textbf{dropout}  & 0\\
 \textbf{input\_normalization}  & False\\
 \textbf{length\_eq}  & 60\\
 \textbf{linear}  & False\\
 \textbf{ln}  & True\\
 \textbf{lr}  & 0.0001\\
 \textbf{mean}  & 0.5\\
 \textbf{n\_l\_enc}  & 5\\
 \textbf{norm}  & True\\
 \textbf{num\_features}  & 20\\
 \textbf{num\_heads}  & 8\\
 \textbf{num\_inds}  & 50\\
 \textbf{output\_dim}  & 60\\
 \textbf{sinuisodal\_embeddings} & False\\
 \textbf{src\_pad\_idx} & 0\\
 \textbf{std} & 0.5\\
 \textbf{trg\_pad\_idx} & 0\\
 \hline
 \end{tabular} 
\label{a-tab6}
}
\end{table*}

\section{Appendix: Detailed Settings of the hyperparameters of MMSR}
\begin{table*}[htbp]
\centering
{
\caption{Hyperparameters of MMSR}
\begin{tabular}{lc}
\textbf{hyperparameters} & \textbf{Numerical value}\\ 
\toprule
 \textbf{$\theta$}  & 0.07\\
 \textbf{$\lambda_{1}$}  & 1.0\\
 \textbf{$\lambda_{2}$}  & 1.0\\
 \textbf{$\lambda_{3}$}  & 1.0\\
 
 \hline
 \end{tabular} 

\label{a-tab6}
}
\end{table*}
\section{Appendix: Test data in detail}
Table \ref{a-tab1},\ref{a-tab2},\ref{a-tab3} shows in detail the expression forms of the data set used in the experiment, as well as the sampling range and sampling number. Some specific presentation rules are described below
\begin{itemize}
\item The variables contained in the regression task are represented as [$x_1,x_2,...,x_n$].
\item $U(a,b,c)$ signifies $c$ random points uniformly sampled between $a$ and $b$ for each input variable. Different random seeds are used for training and testing datasets.
\item $E(a,b,c)$ indicates $c$ points evenly spaced between $a$ and $b$ for each input variable. 
\end{itemize}
\begin{table*}[htbp]
\centering
\caption{ Specific formula form and value range of the three data sets Nguyen, Korns, and Jin. 
}
\begin{scriptsize}
\begin{tabular}{ccccc}
\toprule[1.45pt]
\toprule
Name & Expression & Dataset  \\ \hline
Nguyen-1 & $x_1^3+x_1^2+x_1$&U$(-1, 1, 20)$\\
Nguyen-2 & $x_1^4+x_1^3+x_1^2+x_1$ & U$(-1, 1, 20)$ \\
Nguyen-3 & $x_1^5+x_1^4+x_1^3+x_1^2+x_1$ & U$(-1, 1, 20)$ \\
Nguyen-4 & $x_1^6+x_1^5+x_1^4+x_1^3+x_1^2+x_1$ & U$(-1, 1, 20)$  \\
Nguyen-5 & $\sin(x_1^2)\cos(x)-1$ & U$(-1, 1, 20)$  \\
Nguyen-6 & $\sin(x_1)+\sin(x_1+x_1^2)$ & U$(-1, 1, 20)$  \\
Nguyen-7 & $\log(x_1+1)+\log(x_1^2+1)$ & U$(0, 2, 20)$  \\
Nguyen-8 & $\sqrt{x}$ & U$(0, 4, 20)$  \\
Nguyen-9 & $\sin(x)+\sin(x_2^2)$ & U$(0, 1, 20)$ \\
Nguyen-10 & $2\sin(x)\cos(x_2)$ & U$(0, 1, 20)$ \\
Nguyen-11 & $x_1^{x_2}$ & U$(0, 1, 20)$  \\
Nguyen-12 & $x_1^4-x_1^3+\frac{1}{2}x_2^2-x_2$ & U$(0, 1, 20)$ \\
\toprule
Nguyen-2$'$ & $4x_1^4+3x_1^3+2x_1^2+x$ & U$(-1, 1, 20)$  \\
Nguyen-5$'$ & $\sin(x_1^2)\cos(x)-2$ & U$(-1, 1, 20)$  \\
Nguyen-8$'$ & $\sqrt[3]{x}$ & U$(0, 4, 20)$ \\
Nguyen-8$''$ & $\sqrt[3]{x_1^2}$ & U$(0, 4, 20)$ \\
\toprule
Nguyen-1\textsuperscript{c} & $3.39x_1^3+2.12x_1^2+1.78x$ & U$(-1, 1, 20)$ \\
Nguyen-5\textsuperscript{c} & $\sin(x_1^2)\cos(x)-0.75$ & $U(-1, 1, 20)$  \\
Nguyen-7\textsuperscript{c} & $\log(x+1.4)+\log(x_1^2+1.3)$ & U$(0, 2, 20)$ \\
Nguyen-8\textsuperscript{c} & $\sqrt{1.23 x}$ & U$(0, 4, 20)$  \\
Nguyen-10\textsuperscript{c} & $\sin(1.5x)\cos(0.5x_2)$ & U$(0, 1, 20)$  \\
\toprule
Korns-1 & $1.57+24.3*x_1^4$ & U$(-1, 1, 20)$  \\
Korns-2 & $0.23+14.2\frac{(x_4+x_1)}{(3x_2)}$ & U$(-1, 1, 20)$  \\
Korns-3 & $4.9\frac{(x_2-x_1+\frac{x_1}{x_3}}{(3x_3))}-5.41$ & U$(-1, 1, 20)$ \\
Korns-4 & $0.13sin(x_1)-2.3$ & U$(-1, 1, 20)$  \\
Korns-5 & $3+2.13log(|x_5|)$ & U$(-1, 1, 20)$  \\
Korns-6 & $1.3+0.13\sqrt{|x_1|}$ & U$(-1, 1, 20)$  \\
Korns-7 & $2.1(1-e^{-0.55x_1})$ & U$(-1,1 , 20)$  \\
Korns-8 & $6.87+11\sqrt{|7.23 x_1 x_4 x_5|}$ & U$(-1, 1, 20)$ \\
Korns-9 & $12\sqrt{|4.2x_1x_2x_2|}$ & U$(-1, 1, 20)$ \\
Korns-10 & $0.81+24.3\frac{2x_{1}+3x_2^2}{4x_3^3+5x_4^4}$ & U$(-1, 1, 20)$  \\
Korns-11 & $6.87+11cos(7.23x_1^3)$ & U$(-1, 1, 20)$  \\
Korns-12 & $2-2.1cos(9.8x_1^3)sin(1.3x_5)$ & U$(-1, 1, 20)$  \\ 
Korns-13 & $32.0-3.0\frac{tan(x_1)}{tan(x_2)}\frac{tan(x_3)}{tan(x_4)}$ & U$(-1, 1, 20)$ \\
Korns-14 & $22.0-(4.2cos(x_1)-tan(x_2))\frac{tanh(x_3)}{sin(x_4)}$ & U$(-1, 1, 20)$  \\
Korns-15 & $12.0-\frac{6.0tan(x_1)}{e^{x_2}}(log(x_3)-tan(x_4))))$ & U$(-1, 1, 20)$  \\ 
\toprule
Jin-1 & $2.5 x_1^4-1.3 x_1^3 +0.5 x_2^2 - 1.7x_2$ & U$(-3, 3, 100)$ \\
Jin-2 & $8.0 x_1^2 + 8.0 x_2^3 - 15.0$ & U$(-3, 3, 100)$  \\
Jin-3 & $0.2 x_{1}^{3} + 0.5 x_{2}^{3} - 1.2 x_2 - 0.5 x_{1}$ & U$(-3, 3, 100)$  \\    
Jin-4 & $1.5 \exp{x} + 5.0 cos(x_2)$ & U$(-3, 3, 100)$\\
Jin-5 & $6.0 sin(x_1) cos(x_2)$ & U$(-3, 3, 100)$ \\
Jin-6 & $1.35 x_1 x_2 + 5.5 sin((x_1 - 1.0)(x_2 - 1.0))$ & U$(-3, 3, 100)$ \\   
\toprule
\newline
\end{tabular}
\end{scriptsize}
\label{a-tab1}
\end{table*}

\begin{table*}[htpb]
\centering
\caption{
Specific formula form and value range of the three data sets neat, Keijzer, and Livermore.
}
\begin{scriptsize}
\begin{tabular}{ccccc}
\toprule[1.45pt]
\toprule
Name & Expression & Dataset \\
\hline
Neat-1 & $x_1^4+x_1^3+x_1^2+x$ & U$(-1, 1, 20)$  \\
Neat-2 & $x_1^5+x_1^4+x_1^3+x_1^2+x$ & U$(-1, 1, 20)$ \\
Neat-3 & $\sin(x_1^2)\cos(x)-1$ & U$(-1, 1, 20)$ \\
Neat-4 & $\log(x+1)+\log(x_1^2+1)$ & U$(0, 2, 20)$  \\
Neat-5 & $2\sin(x)\cos(x_2)$ & U$(-1, 1, 100)$  \\
Neat-6 & $\sum_{k=1}^x \frac{1}{k} $ & E$(1, 50, 50)$  \\
Neat-7 & $2 - 2.1\cos(9.8x_1)\sin(1.3x_2)$ & E$(-50, 50, 10^5)$ \\
Neat-8 & $\frac{e^{-(x_1)^2}}{1.2 + (x_2-2.5)^2}$ & U$(0.3, 4, 100)$  \\
Neat-9 & $\frac{1}{1+x_1^{-4}} + \frac{1}{1+x_2^{-4}}$ & E$(-5, 5, 21)$ \\
\toprule
Keijzer-1 & $0.3x_1sin(2\pi x_1)$ & U$(-1, 1, 20)$  \\
Keijzer-2 & $2.0x_1sin(0.5\pi x_1)$ & U$(-1, 1, 20)$  \\
Keijzer-3 & $0.92x_1sin(2.41\pi x_1)$ & U$(-1, 1, 20)$ \\
Keijzer-4 & $x_1^3e^{-x_1}cos(x_1)sin(x_1)sin(x_1)^{2}cos(x_1)-1$ & U$(-1, 1, 20)$ \\
Keijzer-5 & $3+2.13log(|x_5|)$ & U$(-1, 1, 20)$\\

Keijzer-6 & $\frac{x1(x1+1)}{2}$& U$(-1, 1, 20)$ \\
Keijzer-7 & $log(x_1)$ & U$(0,1 , 20)$ \\
Keijzer-8 & $\sqrt{(x_1)}$ & U$(0, 1, 20)$  \\
Keijzer-9 & $log(x_1+\sqrt{x_1^2}+1)$ & U$(-1, 1, 20)$ \\
Keijzer-10 & $x_{1}^{x_2}$ & U$(-1, 1, 20)$  \\
Keijzer-11 & $x_1x_2+sin((x_1-1)(x_2-1))$ & U$(-1, 1, 20)$  \\
Keijzer-12 & $x_1^4-x_1^3+\frac{x_2^2}{2}-x_2$ & U$(-1, 1, 20)$  \\ 
Keijzer-13 & $6sin(x_1)cos(x_2)$ & U$(-1, 1, 20)$  \\
Keijzer-14 & $\frac{8}{2+x_1^2 + x_2^2}$ & U$(-1, 1, 20)$ \\
Keijzer-15 & $\frac{x_1^3}{5}+\frac{x_2^3}{2}-x_2-x_1$ & U$(-1, 1, 20)$ \\ 

\toprule
Livermore-1 & $\frac{1}{3}+x_1+sin(x_1^2))$ & U$(-3, 3, 100)$  \\
Livermore-2 & $sin(x_1^2)*cos(x1)-2$ & U$(-3, 3, 100)$  \\
Livermore-3 & $sin(x_1^3)*cos(x_1^2))-1$ & U$(-3, 3, 100)$  \\
Livermore-4 & $log(x_1+1)+log(x_1^2+1)+log(x_1)$ & U$(-3, 3, 100)$ \\ 
Livermore-5 & $x_1^4-x_1^3+x_2^2-x_2$ & U$(-3, 3, 100)$  \\
Livermore-6 & $4x_1^4+3x_1^3+2x_1^2+x_1$ & U$(-3, 3, 100)$ \\ 
Livermore-7 & $\frac{(exp(x1)-exp(-x_1)}{2})$ & U$(-1, 1, 100)$ \\ 
Livermore-8 & $\frac{(exp(x1)+exp(-x1)}{3}$ & U$(-3, 3, 100)$ \\
Livermore-9 & $x_1^9+x_1^8+x_1^7+x_1^6+x_1^5+x_1^4+x_1^3+x_1^2+x_1$ & U$(-1, 1, 100)$  \\
Livermore-10 & $6*sin(x_1)cos(x_2)$ & U$(-3, 3, 100)$  \\
Livermore-11 & $\frac{x_1^2 x_2^2}{(x_1+x_2)}$ & U$(-3, 3, 100)$ \\
Livermore-12 & $\frac{x_1^5}{x_2^3}$ & U$(-3, 3, 100)$  \\
Livermore-13 & $x_1^{\frac{1}{3}}$ & U$(-3, 3, 100)$  \\
Livermore-14 & $x_1^3+x_1^2+x_1+sin(x_1)+sin(x_2^2)$ & U$(-1, 1, 100)$ \\ 
Livermore-15 & $x_1^\frac{1}{5}$ & U$(-3, 3, 100)$  \\
Livermore-16 & $x_1^{\frac{2}{3}}$ & U$(-3, 3, 100)$  \\  
Livermore-17 & $4sin(x_1)cos(x_2)$ & U$(-3, 3, 100)$  \\
Livermore-18 & $sin(x_1^2)*cos(x_1)-5$ & U$(-3, 3, 100)$  \\
Livermore-19 & $x_1^5+x_1^4+x_1^2 + x_1$ & U$(-3, 3, 100)$  \\
Livermore-20 & $e^{(-x_1^2)}$ & U$(-3, 3, 100)$  \\
Livermore-21 & $x_1^8+x_1^7+x_1^6+x_1^5+x_1^4+x_1^3+x_1^2+x_1$& U$(-1, 1, 20)$ \\
Livermore-22 & $e^{(-0.5x_1^2)}$ & U$(-3, 3, 100)$  \\
\toprule
\newline
\end{tabular}
\end{scriptsize}
\label{a-tab2}
\end{table*}

\begin{table*}[htpb]
\centering
\caption{
Specific formula form and value range of the three data sets Vladislavleva and others. }
\begin{scriptsize}
\begin{tabular}{ccccc}
\toprule[1.45pt]
\toprule
Name & Expression & Dataset \\
\toprule
Vladislavleva-1 & $\frac{(e^{-(x1-1)^2})}{(1.2+(x2-2.5)^2))}$ & U$(-1, 1, 20)$ \\
Vladislavleva-2 & $e^{-x_1}x_1^3cos(x_1)sin(x_1)(cos(x_1)sin(x_1)^2-1)$ & U$(-1, 1, 20)$ \\

Vladislavleva-3 & $e^{-x_1}x_1^3cos(x_1)sin(x_1)(cos(x_1)sin(x_1)^2-1)(x_2-5)$ & U$(-1, 1, 20)$ \\
Vladislavleva-4 & $\frac{10}{5+(x1-3)^2+(x_2-3)^2+(x_3-3)^2+(x_4-3)^2+(x_5-3)^2}$ & U$(0, 2, 20)$ \\
Vladislavleva-5 & $30(x_1-1)\frac{x_3-1}{(x_1-10)}x_2^2$ & U$(-1, 1, 100)$ \\
Vladislavleva-6 & $6sin(x_1)cos(x_2)$ & E$(1, 50, 50)$ \\
Vladislavleva-7 & $2 - 2.1\cos(9.8x)\sin(1.3x_2)$ & E$(-50, 50, 10^5)$ \\
Vladislavleva-8 & $\frac{e^{-(x-1)^2}}{1.2 + (x_2-2.5)^2}$ & U$(0.3, 4, 100)$  \\
\toprule
Test-2 & $3.14x_1^2$ & U$(-1, 1, 20)$ \\
Const-Test-1 & $5x_1^2$ & U$(-1, 1, 20)$ \\
GrammarVAE-1 & $1/3+x1+sin(x_1^2))$ & U$(-1, 1, 20)$ \\
Sine & $sin(x_1)+sin(x_1+x_1^2))$ & U$(-1, 1, 20)$ \\
Nonic & $x_1^9+x_1^8+x_1^7+x_1^6+x_1^5+x_1^4+x_1^3+x_1^2+x_1$ & U$(-1, 1, 100)$  \\
Pagie-1 & $\frac{1}{1+x_1^{-4}+\frac{1}{1+x2^{-4}}} $ & E$(1, 50, 50)$  \\
Meier-3 & $\frac{x_1^2  x_2^2}{(x_1+x_2)}$ & E$(-50, 50, 10^5)$ \\
Meier-4 & $\frac{x_1^5}{x_2^3}$ & $U(0.3, 4, 100)$  \\
Poly-10 & $x_1x_2+x_3x4+x_5x_6+x_1x_7x_9+x_3x_6x_{10}$ & E$(-1, 1, 100)$ \\
\toprule
Constant-1 & $3.39*x_1^3+2.12*x_1^2+1.78*x_1$&$U(-4, 4, 100)$\\
Constant-2 & $sin(x_1^2)*cos(x_1)-0.75$&$U(-4, 4, 100)$\\
Constant-3 & $sin(1.5*x_1)*cos(0.5*x_2)$&$U(0.1, 4, 100)$\\
Constant-4 & $2.7*x_1^{x_2}$&$U(0.3, 4, 100)$\\
Constant-5 & $sqrt(1.23*x_1)$&$U(0.1, 4, 100)$\\
Constant-6 & $x_1^{0.426}$&$U(0.0, 4, 100)$\\
Constant-7 & $2*sin(1.3*x_1)*cos(x_2)$&$U(-4, 4, 100)$\\
Constant-8 & $log(x_1+1.4)+log(x1,2+1.3)$&$U(-4, 4, 100)$\\
\toprule
R1 & $\frac{(x_1+1)^3}{x_1^2-x_1+1)}$&$U(-5, 5, 100)$\\
R2 & $\frac{(x_1^2-3*x_1^2+1}{x_1^2+1)}$&$U(-4, 4, 100)$\\
R3 & $\frac{x_1^6+x_1^5)}{(x_1^4+x_1^3+x_1^2+x1+1)}$&$U(-4, 4, 100)$\\
\toprule
\newline
\end{tabular}
\end{scriptsize}
\label{a-tab3}
\end{table*}

\section{Appendix: MMSR tests on AIFeynman dataset.}

In our study, we conducted an evaluation of our novel symbol regression algorithm, termed MMSR, leveraging the AI Feynman dataset, which comprises a diverse array of problems spanning various subfields of physics and mathematics, including mechanics, thermodynamics, and electromagnetism. Originally, the dataset contains 100,000 data points; however, for a more rigorous assessment of MMSR's efficacy, our analysis was deliberately confined to a subset of 100 data points. Through the application of MMSR for symbol regression on these selected data points, we meticulously calculated the $R^2$ values to compare the algorithm's predictions against the true solutions.

The empirical results from our investigation unequivocally affirm that MMSR possesses an exceptional ability to discern the underlying mathematical expressions from a constrained sample size. Notably, the $R^2$ values achieved were above 0.99 for a predominant portion of the equations, underscoring the algorithm's remarkable accuracy in fitting these expressions. These findings decisively position MMSR as a potent tool for addressing complex problems within the domains of physics and mathematics. The broader implications of our study suggest that MMSR holds considerable promise for a wide range of applications across different fields. Detailed experimental results are presented in Table \ref{a-tab5} and Table \ref{a-tab6}.

\begin{table}[htbp]
\centering
{\footnotesize
\begin{tabular}{|l|l|r|}
\hline
Feynman   & Equation & $R^2$ \\
\hline                            
I.6.20a       & $f = e^{-\theta^2/2}/\sqrt{2\pi}$ & 0.9998  \\
I.6.20        & $f = e^{-\frac{\theta^2}{2\sigma^2}}/\sqrt{2\pi\sigma^2}$ & 0.9982\\
I.6.20b       & $f = e^{-\frac{(\theta-\theta_1)^2}{2\sigma^2}}/\sqrt{2\pi\sigma^2}$ & 0.9901 \\
I.8.14       & $d = \sqrt{(x_2-x_1)^2+(y_2-y_1)^2}$ & 0.9213  \\
I.9.18       & $F = \frac{Gm_1m_2}{(x_2-x_1)^2+(y_2-y_1)^2+(z_2-z_1)^2}$  & 0.9928\\
I.10.7       & $F = \frac{Gm_1m_2}{(x_2-x_1)^2+(y_2-y_1)^2+(z_2-z_1)^2}$  & 0.9901\\
I.11.19      & $A = x_1y_1+x_2y_2+x_3y_3$ & 0.9999   \\
I.12.1       & $F = \mu N_n$ & 0.9998 \\
I.12.2       & $F = \frac{q_1q_2}{4\pi\epsilon r^2}$   & 1.0 \\
I.12.4       & $E_f = \frac{q_1}{4\pi\epsilon r^2}$  & 0.9992 \\
I.12.5       & $F = q_2 E_f$ & 1.0  \\
I.12.11      & $F = \mathcal{Q}(E_f+B v \sin\theta)$  & 0.9989 \\
I.13.4      & $K = \frac{1}{2}m(v^2+u^2+w^2)$  & 0.9991  \\
I.13.12      & $U = Gm_1m_2(\frac{1}{r_2}-\frac{1}{r_1})$ & 0.9993  \\
I.14.3       & $U = mgz$ &0.9999    \\
I.14.4       & $U = \frac{k_{spring}x^2}{2}$  & 0.9903  \\
I.15.3x      & $x_1 = \frac{x-ut}{\sqrt{1-u^2/c^2}}$ & 0.9802 \\
I.15.3t      & $t_1 = \frac{t-ux/c^2}{\sqrt{1-u^2/c^2}}$ & 0.9729  \\
I.15.10       & $p = \frac{m_0v}{\sqrt{1-v^2/c^2}}$ & 0.99983 \\
I.16.6       & $v_1 = \frac{u+v}{1+uv/c^2}$ & 0.9873  \\
I.18.4       & $r = \frac{m_1r_1+m_2r_2}{m_1+m_2}$ & 0.9819 \\
I.18.12      & $\tau = rF\sin\theta$  & 0.9995  \\
I.18.16      & $L = mrv \sin\theta$  & 0.9939 \\
I.24.6 & $E = \frac{1}{4} m (\omega^2+\omega_0^2) x^2$      & 0.9971\\
I.25.13      & $V_e = \frac{q}{C}$ & 1.0 \\
I.26.2       & $\theta_1 = \arcsin(n  \sin\theta_2)$ & 0.9983 \\
I.27.6       & $f_f$    $ = \frac{1}{\frac{1}{d_1}+\frac{n}{d_2}}$  & 0.9992 \\
I.29.4       & $k = \frac{\omega}{c}$ & 0.9998 \\
I.29.16      & $x = \sqrt{x_1^2+x_2^2-2x_1x_2\cos(\theta_1-\theta_2)}$ & 0.9923  \\
I.30.3 & $I_* = I_{*_0}\frac{\sin^2(n\theta/2)}{\sin^2(\theta/2)}$ & 0.9936 \\
I.30.5       & $\theta = \arcsin(\frac{\lambda}{nd})$  & 0.9926\\
I.32.5       & $P = \frac{q^2a^2}{6\pi\epsilon c^3}$       & 0.9992 \\
I.32.17 & $P = (\frac{1}{2}\epsilon c E_f^2)(8\pi r^2/3) (\omega^4/(\omega^2-\omega_0^2)^2)$      & 0.9849  \\
I.34.8       & $\omega = \frac{qvB}{p}$   & 0.9999\\
I.34.10       & $\omega = \frac{\omega_0}{1-v/c}$ & 0.9915 \\
I.34.14      & $\omega = \frac{1+v/c}{\sqrt{1-v^2/c^2}}\omega_0$  & 0.9982 \\
I.34.27      & $E = \hbar\omega$  & 0.9999 \\
I.37.4       & $I_* = I_1+I_2+2\sqrt{I_1I_2}\cos\delta$ & 0.9961\\
I.38.12      & $r = \frac{4\pi\epsilon\hbar^2}{mq^2}$   & 0.9999  \\
I.39.10       & $E = \frac{3}{2}p_F V$     & 0.9999 \\
I.39.11      & $E = \frac{1}{\gamma-1}p_F V$  & 0.9993 \\
I.39.22      & $P_F = \frac{n k_b T}{V}$       & 0.9981  \\
I.40.1       & $n = n_0e^{-\frac{mgx}{k_bT}}$    & 0.9868 \\
I.41.16      & $L_{rad} = \frac{\hbar\omega^3}{\pi^2c^2(e^{\frac{\hbar\omega}{k_bT}}-1)}$ & 0.9453  \\
I.43.16      & $v = \frac{\mu_{drift}q V_e}{d}$   & 0.9941  \\
I.43.31      & $D = \mu_e k_bT$    & 1.0  \\
I.43.43      & $\kappa = \frac{1}{\gamma-1}\frac{k_bv}
{A}$  & 0.9513  \\
I.44.4       & $E = n k_b T \ln(\frac{V_2}{V_1})$   & 0.8616  \\
I.47.23      & $c = \sqrt{\frac{\gamma pr}{\rho}}$   & 0.9639\\
I.48.20       & $E = \frac{m c^2}{\sqrt{1-v^2/c^2}}$ &  0.8935\\
I.50.26 & $x = x_1[\cos(\omega t)+\alpha\> cos(\omega t)^2]$      & 0.9999   \\
\hline
\end{tabular}
\caption{Tested Feynman Equations, part 1.}
\label{a-tab5}
}
\end{table}
\begin{table*}[htbp]
\centering
{\footnotesize

\begin{tabular}{|l|l|r|}
\hline
Feynman   & Equation & $R^2$\\
\hline       
II.2.42   & P     $ = \frac{\kappa(T_2-T_1)A}{d}$  & 0.8511  \\
II.3.24   & $F_E = \frac{P}{4\pi r^2}$  & 0.9910 \\
II.4.23   & $V_e = \frac{q}{4\pi\epsilon r}$   & 0.9892 \\
II.6.11 & $V_e =\frac{1}{4\pi\epsilon}\frac{p_d\cos \theta}{r^2}$      & 0.9939 \\
II.6.15a & $E_f = \frac{3}{4\pi\epsilon}\frac{p_d z}{r^5} \sqrt{x^2+y^2}$      & 0.9650  \\
II.6.15b & $E_f = \frac{3}{4\pi\epsilon}\frac{p_d}{r^3} \cos\theta\sin\theta$      & 0.9928  \\
II.8.7    & $E = \frac{3}{5}\frac{q^2}{4\pi\epsilon d}$  & 0.9906  \\
II.8.31   & $E_{den} = \frac{\epsilon E_f^2}{2}$                     & 0.9999 \\
II.10.9   & $E_f = \frac{\sigma_{den}}{\epsilon}\frac{1}{1+\chi}$      & 0.9995  \\
II.11.3 & $x = \frac{q E_f}{m(\omega_0^2-\omega^2)}$      & 0.9904     \\
II.11.7 & $n = n_0(1+ \frac{p_d E_f \cos\theta}{k_b T})$      & 0.8918 \\
II.11.20  & $P_* = \frac{n_\rho p_d^2 E_f}{3 k_b T}$ & 0.8311  \\
II.11.27 & $P_* = \frac{n\alpha}{1-n\alpha/3}\epsilon E_f$      & 0.9955   \\
II.11.28  & $\theta = 1+\frac{n\alpha}{1-(n\alpha/3)}$    & 0.9999\\ 
II.13.17  & $B = \frac{1}{4 \pi \epsilon c^2}\frac{2I}{r}$ & 0.9853\\
II.13.23  & $\rho_c = \frac{\rho_{c_0}}{\sqrt{1-v^2/c^2}}$          & 0.9723  \\
II.13.34  & $j = \frac{\rho_{c_0}v}{\sqrt{1-v^2/c^2}}$     & 0.9907 \\
II.15.4   & $E = -\mu_M B \cos\theta$               & 0.9999 \\
II.15.5   & $E = -p_d E_f\cos\theta$  & 0.9999 \\
II.21.32  & $V_e = \frac{q}{4\pi\epsilon r(1-v/c)}$   & 0.9983   \\
II.24.17 & $k = \sqrt{\frac{\omega^2}{c^2}-\frac{\pi^2}{d^2}}$      & 0.9952   \\
II.27.16  & $F_E = \epsilon c E_f^2$        & 0.9991 \\
II.27.18  & $E_{den} = \epsilon E_f^2$         & 0.9989 \\
II.34.2a  & $I = \frac{qv}{2\pi r}$         & 0.9938 \\
II.34.2   & $\mu_M = \frac{q v r}{2}$             & 0.9935 \\
II.34.11  & $\omega = \frac{g_{\_} q B}{2m}$          & 0.9903 \\
II.34.29a & $\mu_M = \frac{q h}{4\pi m}$      & 0.9830  \\
II.34.29b & $E = \frac{g_{\_} \mu_M B J_z}{\hbar}$ & 0.8013\\
II.35.18 & $n = \frac{n_0}{\exp(\mu_m B/(k_b T))+\exp(-\mu_m B/(k_b T))}$      & 0.9244 \\
II.35.21  & $M = n_\rho \mu_M \tanh(\frac{\mu_M B}{k_b T})$     & 0.7822 \\
II.36.38 & $f = \frac{\mu_m B}{k_b T}+\frac{\mu_m\alpha M}{\epsilon c^2 k_b T}$      & 0.9550\\
II.37.1   & $E = \mu_M(1+\chi)B$    & 0.9924\\
II.38.3   & $F = \frac{Y A x}{d}$            & 0.9999 \\
II.38.14  & $\mu_S = \frac{Y}{2(1+\sigma)}$     & 0.9994  \\
III.4.32  & $n = \frac{1}{e^{\frac{\hbar\omega}{k_bT}}-1}$ & 0.9322  \\
III.4.33  & $E = \frac{\hbar\omega}{e^{\frac{\hbar\omega}{k_b T}}-1}$  & 0.9976    \\
III.7.38  & $\omega = \frac{2 \mu_M B}{\hbar}$  & 0.9922  \\
III.8.54  & $p_{\gamma}$    $ = \sin(\frac{E t}{\hbar})^2$  & 0.9723\\
III.9.52  & $p_{\gamma}$    $ = \frac{p_d E_f t}{\hbar} \frac{    \sin((\omega-\omega_0)t/2)^2}{((\omega-\omega_0)t/2)^2}$ & 0.7323  \\
III.10.19 & $E = \mu_M\sqrt{B_x^2+B_y^2+B_z^2}$  & 0.9838 \\
III.12.43 & $L = n\hbar$ & 0.9794  \\
III.13.18 & $v = \frac{2 E d^2 k}{\hbar}$ & 0.9989  \\
III.14.14 & $I = I_0 (e^{\frac{q V_e}{k_b T}}-1)$  & 0.9825\\
III.15.12 & $E = 2U(1-\cos(kd))$    & 0.9999 \\
III.15.14 & $m = \frac{\hbar^2}{2E d^2}$     & 0.9990  \\
III.15.27 & $k = \frac{2\pi\alpha}{nd}$    & 0.9982 \\
III.17.37 & $f = \beta(1+\alpha \cos\theta)$ & 0.9413 \\
III.19.51 & $E = \frac{-mq^4}{2(4\pi\epsilon)^2\hbar^2}\frac{1}{n^2}$     & 0.9739 \\
III.21.20 & $j = \frac{-\rho_{c_0} q A_{vec}}{m}$  & 0.8113  \\
\hline
\end{tabular} 
\caption{Tested Feynman Equations, part 2.}
\label{a-tab6}
}
\end{table*}

\end{document}